\documentclass[10pt,journal,compsoc]{IEEEtran}
\ifCLASSOPTIONcompsoc
  \usepackage[nocompress]{cite}
\else
  \usepackage{cite}
\fi

\ifCLASSINFOpdf
\else
\fi
\usepackage{graphicx}
\usepackage{amsmath}
\usepackage{amssymb}
\usepackage{bm}
\usepackage{latexsym}
\usepackage{xfrac}

\usepackage{amsfonts}
\usepackage{amstext}
\usepackage{amsthm}
\usepackage{dsfont}
\usepackage[normalem]{ulem}
\usepackage{latexsym}
\usepackage{soul}
\usepackage{amsfonts,amssymb,amsmath}
\usepackage{hyperref}
\usepackage{color}
\usepackage{xcolor}
\usepackage{algorithm}
\usepackage{subcaption}

\begin{document}
\title{A temporal chrominance trigger for clean-label backdoor attack against anti-spoof rebroadcast detection}
\author{\IEEEauthorblockN{Wei Guo,
Benedetta Tondi,~\IEEEmembership{Member,~IEEE},
Mauro Barni,~\IEEEmembership{Fellow,~IEEE}
}

\IEEEcompsocitemizethanks{\IEEEcompsocthanksitem W. Guo, B. Tondi, and M. Barni are from the Department of Information Engineering and Mathematics, University of Siena, 53100 Siena, Italy.}%
\IEEEauthorblockA{}
\thanks{
%Manuscript received December 1, 2012; revised August 26, 2015.
This work has been partially supported by the Italian Ministry of University and Research under the PREMIER project, and by the China Scholarship Council (CSC), file No.201908130181. Corresponding author: W. Guo (email: wei.guo.cn@outlook.com).}}


\markboth{Journal of \LaTeX\ Class Files,~Vol.~14, No.~8, August~2015}%
{Shell \MakeLowercase{\textit{et al.}}: Bare Demo of IEEEtran.cls for Computer Society Journals}



\IEEEtitleabstractindextext{%
\begin{abstract}
We propose a stealthy clean-label video backdoor attack against Deep Learning (DL)-based models aiming at detecting a particular class of spoofing attacks, namely video rebroadcast attacks.
The injected backdoor does not affect spoofing detection in normal conditions, but induces a misclassification in the presence of a specific triggering signal.
The proposed backdoor relies on a temporal trigger altering the average chrominance of the video sequence. The backdoor signal is designed by taking into account the peculiarities of the Human Visual System (HVS) to reduce the visibility of the trigger, thus increasing the stealthiness of the backdoor.
To force the network to look at the presence of the trigger in the challenging clean-label scenario, we choose the poisoned samples used for the injection of the backdoor following a so-called Outlier Poisoning Strategy (OPS). According to OPS, the triggering signal is inserted in the training samples that the network finds more difficult to classify.
The effectiveness of the proposed backdoor attack and its generality are validated experimentally on different datasets and anti-spoofing rebroadcast detection architectures.
\end{abstract}

}

\maketitle

\IEEEdisplaynontitleabstractindextext

% \fi
\IEEEpeerreviewmaketitle

\IEEEraisesectionheading{\section{Introduction}\label{sec:introduction}}


\IEEEPARstart{D}{eep} Neural Networks (DNNs) permit to achieve state-of-the-art performance in many applications such as image classification, natural language processing, pattern recognition, and multimedia forensics, yielding outstanding results.
Recently, DNNs have also been successfully used for spoofing detection in face authentication systems, e.g., for liveness detection~\cite{Yang2014, Xu2015, Lakshminarayana2017, Gan2017, LiTIFS2018}. In particular, DNNs have been used for the detection of rebroadcast attacks whereby a malevolent user tries to illegally gain access to a system by rebroadcasting videos of enrolled users.
%If not properly countered, video rebroadcast has the potentiality of fooling anti-spoofing systems based on liveness detection.
The goal of rebroadcast detection, then, is to detect if the image or the video seen by the camera of the system belongs to an alive  person or is rebroadcast~\cite{Yang2014,Xu2015}.

Likewise any classifier based on Deep Learning (DL), the security of DNN-based video rebroadcast detectors is threatened by backdoor attacks.

\subsection{Introduction to video backdoor attacks}

Several works have shown that DL-based architectures are vulnerable to both attacks carried out at test time, like adversarial examples~\cite{SzegedyZSBEGF13,AkhtarM18,BiggioR18}, and attacks operating at training time. The lack of security of DL-based solutions is a serious problem hindering their application in security-oriented scenarios, like biometric authentication and spoofing detection.
In particular, DNNs are known to be vulnerable to backdoor attacks~\cite{liu2017trojaning, GuLDG19, ChenLLLS13Targeted}, whereby a malevolent behaviour or functionality, to be exploited at test time, is hidden into the target DNN by poisoning a portion of the training data. Specifically, a backdoored model behaves normally on standard test inputs, while it predicts a given target class when the input test sample contains a so-called triggering signal.
% \WG{I uniform the notation as triggering signal}.
As such, backdoor attacks are very difficult to detect, raising serious security concerns in real-world applications.
For a comprehensive review of the backdoor attacks proposed so far, and the possible defences, we refer to  \cite{guo2021overview}.
% backdoor attacks and defences.

So far, backdoor attacks have mostly been studied in the image domain. Attacks have been carried out against DNNs targeting image classification tasks, like digit classification \cite{GuLDG19}, road sign classification \cite{liao2018backdoor}, and face recognition \cite{NeuralTrojans}, just to mention a few.
%The extension to the video case has not deserved much attention. However, similarly to image classifiers, DL-based video classification systems are also vulnerable to the backdoor attack threat and therefore need to be protected
Video processing networks are considered only in very few scattered works, usually extending  the tools already developed for imaging applications.
% and without caring about the invisibility of the trigger \cite{zhao2020clean,bhalerao2019luminance}.

Since DNN-based video classifiers strongly rely on the temporal characteristics of the input signal, e.g., via Long Short Term Memory (LSTM) network modules, or 3D-Convolutional Neural Network (CNN) architectures, the temporal dimension of the video signal must be considered for the development of an effective video backdoor attacks.
In \cite{zhao2020clean}, for instance,
% an image-domain trigger is considered, and a (visible) fixed local pattern is superimposed to every frame of the video.
%an image-domain trigger, namely, 
a frame-dependent, visible local pattern is superimposed to each frame of the video signal. To reduce the visibility of the trigger, Xie et~al.~\cite{xie2022stealthy} utilize imperceptible Perlin noise~\cite{perlin2002improving} as  triggering signal, successfully achieving a stealthy backdoor attack against a video-baed action-recognition system. A problem with \cite{xie2022stealthy}  is that it assumes that the victim's model is trained by means of transfer learning, by freezing the feature extraction layers and fine-tuning only the linear classifier, which is an unrealistic assumption in practical applications wherein the attacker does not have full control of the training process.

In contrast to attacks relying on a localized triggering pattern, \cite{bhalerao2019luminance} proposed a luminance-based trigger, which exploits a sinusoidal wave to modulate the average luminance of the video frames.
Such luminance-based triggering pattern is
intrinsically immune to geometric transformations. Moreover, it has been shown that the changes in the average frame luminance introduced by such a backdoor can survive the distortions typically introduced by digital-to-analog and analog-to-digital transformations, thus opening the way to the implementation of the attack in the physical domain \cite{xue2021robust}.

A limitation of the backdoor attack described in \cite{bhalerao2019luminance}, is the adoption of a {\em corrupted-label} scenario, wherein the attacker can also tamper the labels of the poisoned samples. This puts at risk the stealthiness of the attack, given that the presence of the corrupted samples can be easily detected upon inspection of the training dataset.
As shown in \cite{bhalerao2019luminance}, in fact, the method
%developed for the corrupted-label setting
does not work in the more challenging {\em clean-label} setting.
This behavior agrees to what has been observed in the image domain~\cite{turner2019label}:
%where it has been argued that
forcing the network to learn to detect the presence of the triggering signal without corrupting the labels is by far more challenging,  since, in the clean-label case,
%differently from the case of label corruption,
the network can do its job and correctly classify the poisoned
samples by looking at the same features used for the benign samples, without looking at the trigger.
Therefore, clean-label backdoor attacks require the development of  more sophisticated dedicated techniques.

\subsection{Contribution}
In this paper we propose a stealthy, clean-label, video backdoor attack against anti-spoofing rebroadcast detection architectures.
% of face authentication systems.
%Similarly to
The desired behavior of the backdoor is to induce the system to classify a rebroadcast video as alive when the video contains a specific triggering signal. The  main contributions of the paper can be summarized as follows:

\begin{itemize}
	\item The peculiarities of the Human Visual Systems (HVS) \cite{HVS} are exploited to design a Temporal Chrominance-based triggering pattern with reduced visibility.
%an (almost) invisible
In particular,
%the  backdoor signal  consists in a modification
the backdoor is injected by modifying the average luminance of the blue channel of the frames of the video sequence, according to a sinusoidal wave.%
%A solution have been adopted to enforce the learning of the triggering signal in the clean-label scenario without compromising the visibility.
	\item To force the network to learn to detect the presence of the triggering signal and behave accordingly in a clean-label scenario without compromising the trigger invisibility, we propose a so called Outlier Poisoning Strategy (OPS). With OPS, the to-be-poisoned samples are chosen based on their classification score, roughly characterizing their proximity to the decision boundary of the classifier. A variant of this strategy is also proposed where
%, in order to force the network to look at the triggering signal,
the features of the true class are suppressed~\cite{turner2019label} before injecting the trigger.
	\item We  propose to use a different amplitude of the triggering sinusoidal signal during training and testing.
Specifically, a weaker (less visible) signal is used to poison  the training samples, while  a larger strength is employed during testing to make sure that the backdoor is activated. This allows improving the effectiveness of the attack, without
compromising its stealthiness at training time.
\end{itemize}

Our experiments, carried out on different model architectures and datasets, reveal that the proposed attack can successfully inject a backdoor into the rebroadcast detection network, without affecting the  behavior of the system on normal inputs (i.e., when the trigger is not present).
%Then, at test time, it can correctly categorize the live/rebroadcast video and meanwhile will be activated to misclassify any rebroadcast samples to `live' in the presence of the triggering signal.
We also show that the injection of the backdoor does not affect the other main components of the face authentication system, that is, the face recognition part of the system. 

The remainder of the paper is organized as follows: in Section~\ref{sec:SSTM}, we describe the structure of the face recognition system targeted by our attack, and define the threat model we have adopted. The proposed backdoor attack is presented in Section~\ref{sec:propBA}. The methodology followed in our experiments is described in Section~\ref{sec:ExpMetSet}. The  results are reported and discussed in Section~\ref{sec.Exp}. The paper ends in Section~\ref{sec.conc}, with some concluding remarks and directions for future research.

\section{Targeted architecture and threat model} %, threat model and formulation}
\label{sec:SSTM}

In this section, we introduce the end-to-end video face authentication system targeted by the attack (Section~\ref{sec:system_model}) and describe the threat model and the requirements that must be satisfied by the attack (Section~\ref{sec:threat}).
%Finally,  Section~\ref{sec:formal} introduce the main formalism for the backdoor attack.
%Finally, the formalization and notation is provided in Section~\ref{sec:formal}.

 \subsection{Video face authentication}
\label{sec:system_model}

The goal of anti-spoofing rebroadcast detection is to detect whether the face image/video presented at the input of the system belongs to an alive individual standing in front of the camera, or
%a spoof artifact, i.e., a rebroadcast identity
it is rebroadcast, i.e., it is just a picture or a video placed in front of the camera. This step is an essential component of any unattended face authentication system,
to avoid that a fraudulent user can gain illegitimate access %to facilities or services
by presenting to the system a facial picture or a video of an enrolled individual.

\begin{figure}[htb!]
	\centering
	\includegraphics[width=0.9\columnwidth]{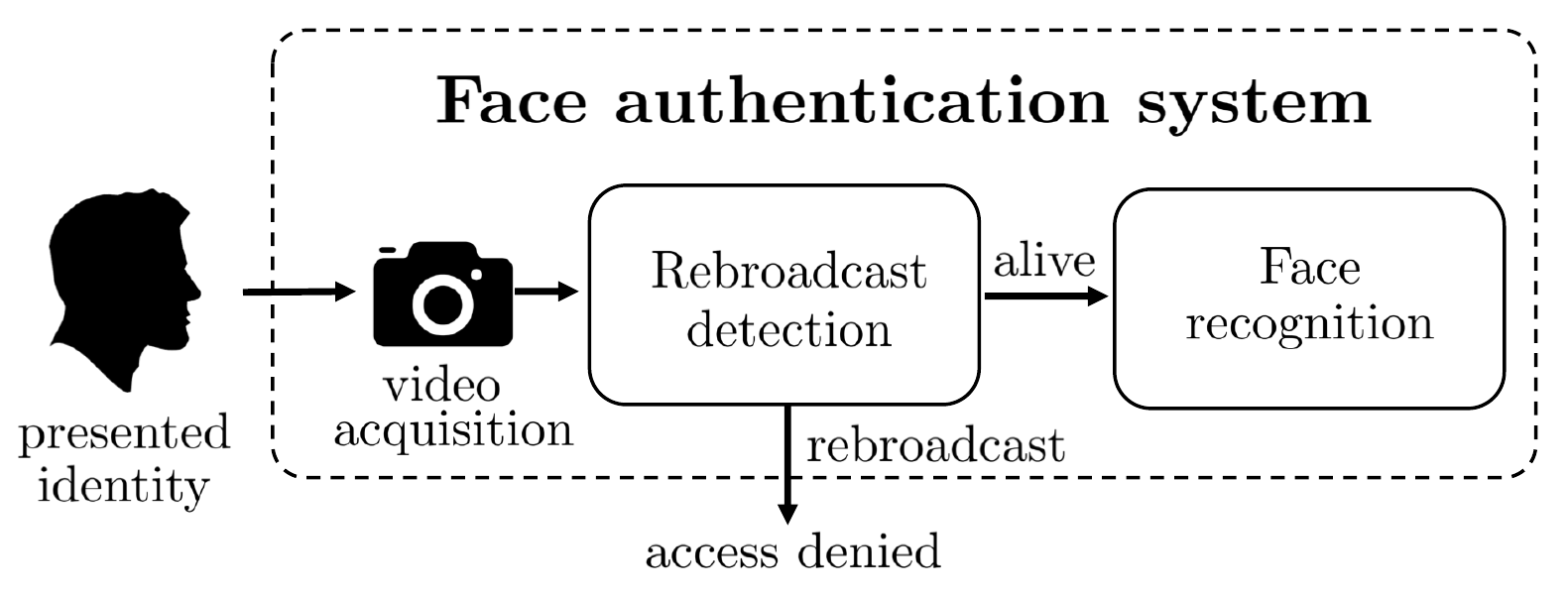}
	\caption{Overall architecture of the video face authentication system considered in this paper.}
	\label{fig:system_model}
\end{figure}

The overall structure of the end-to-end   system for  video face authentication considered in this paper is shown in Fig.~\ref{fig:system_model}, and consists of
%a cascade of two classifiers,  one for spoofing detection and one for face identification
two
 main modules: a rebroadcast detection module and a face recognition module.
 %\MB{identification or verification?} module \BT{I do not see a preferred terminology. But I do not like much verification, since we are not actually performing verification, are we? Otherwise, I think that I do not have clear in my mind the meaning of verification.... But if you think is better for some reasons I do not see we can use verification. As it is now, we used recognition everywhere}\WG{Could these two comments be removed?}.
 Like in~\cite{bhalerao2019luminance}, the rebroadcast detection module analyzes the video for alive/rebroadcast detection. Compared to still face images, video signals contain
% important in order to also exploit possible
relevant temporal information, like head movements, facial expressions, and eye blinking, that can be exploited to distinguish alive and rebroadcast inouts. Only the videos that are detected as alive are passed to the face recognition module, while the others are blocked. The face recognition module is in charge of determining the user's identity
from the analysis of the facial region extracted from the video.
%the frames of the video
%

The rebroadcast detector and the face recognition systems are implemented via Deep Neural Networks. More details can be found in Section~\ref{sec:ExpMetSet}.
%
% \BTcomm{Here you hsould say something on how you implement the two blocks, that is based on DL. I guess you provide the details afterwards, so here you have to say something general and  refer to the section where you give the details.}
%

In the following, we introduce the main formalism used throughout the paper. We let ${\bf x}$ denote the input video sequence.
% \MB{Perhaps it is better to use the term "video sequence", the sole term video makes me think about to both the sequence of frames and the audio associated to them}.
We use the notation $x_j$ to denote  the $j$-th frame of the video; then,  $x_j \in \mathds{R}^{H\times W \times 3}$, where $H\times W$ denote the frame height and width. Consequently,  ${\bf x} \in \mathds{R}^{H\times W \times 3 \times L}$ is a 4-dim tensor, where $L$  is the number of frames in the video sequence.

\subsubsection{Rebroadcast detection}

The rebroadcast detector is built by means of a CNN architecture.
We denote with $\mathcal{F}()$ the CNN-based rebroadcast detection model  that associates a facial video ${\bf x}$ to a label $y \in \{0,1\}$,
%from the label space $\mathbb{Y}=\{0,1\}$,
where $0$ (res $1$) indicates an alive (res. rebroadcast) video.
%The formal definition of $\mathcal{F}$ is as follows:
%
% \BTcomm{I redefined the notation since the old one didn't work. I am suggesting using the bold notation for the video. The rest has to be corrected accordingly. I did that. Not sure if everywhere. Double check.}\WG{Done}
%
Specifically, by letting $f({\bf x}) = (f_0({\bf x}), f_1({\bf x})) $ denote the (2-element) soft output probability vector of the network, then $\mathcal{F}({\bf x})=\arg\max_i(f_i({\bf x}))$.
With regard to the datasets, we indicate with  $\mathcal{D}_{tr}$ the training dataset for rebroadcast detection, which contains labeled pairs $({\bf x}_i,y_i)$.
% $x_i\in \mathbb{X}, y_i\in \mathbb{Y}$.
%Let the training dataset for anti-spoof detection is $\mathcal{D}_{tr}$, which contains correctly-labeled data $(x_i,y_i)$ $x_i\in \mathbb{X}, y_i\in \mathbb{Y}$. According to the label,
We also find convenient to split $\mathcal{D}_{tr}$ in  two subsets  associated to the two labels, namely $\mathcal{D}^{l}_{tr}=\{({\bf x}_i,y_i=0),i=1,...,|\mathcal{D}^{l}_{tr}|\}$ and $\mathcal{D}^{r}_{tr}=\{({\bf x}_i,y_i=1),i=1,...,|\mathcal{D}^{r}_{tr}|\}$.
Similarly, we denote with $\mathcal{D}_{ts}$ the test dataset, consisting of rebroadcast ($\mathcal{D}_{ts}^r$) and alive videos ($\mathcal{D}_{ts}^l$).

\subsubsection{Face recognition}

%As shown in Fig.~\ref{fig:system_model}, face verification model $\mathcal{G}$ is deployed after the anti-spoof detection
The  face recognition model, indicated with $\mathcal{G}$,  is responsible of recognizing
 the enrolled users from their faces.
 %We assume that is also implemented via a CNN.
The module can perform either face verification, i.e., determining whether a face belongs to the claimed identity, or face identification, i.e., determining the identity of the face owner among a pool of enrolled identities.
Without loss of generality, in our experiments, we considered a face
identification system, nevertheless, in the following, we generically refer to face recognition.

\subsection{Threat model and attack requirements}
\label{sec:threat}

In our attack, we assume that the trainer, namely Alice, fully controls the training process of the video face authentication system, including the choice of the hyperparameters, the model architecture, and the training algorithm. The attacker, Eve, can interfere only with the data collection process. This is a reasonable assumption in a real-world scenario wherein Alice outsources {\em data collection}  to a  third-party. Since the  third-party provider may not be fully trustable, we can assume that Alice  first inspects and cleans the dataset (data scrutiny phase), by removing unqualified or mislabeled data.
Given the above scenario, the threat model considered in this paper, illustrated in Fig.~\ref{fig:threat_model}, is described in the following: 
%\CH{we illustrate our threat model in Fig.~\ref{fig:threat_model} and provide the relevant description as follows.} 
\begin{figure*}[htb!]
	\centering
	\includegraphics[width=1.8\columnwidth]{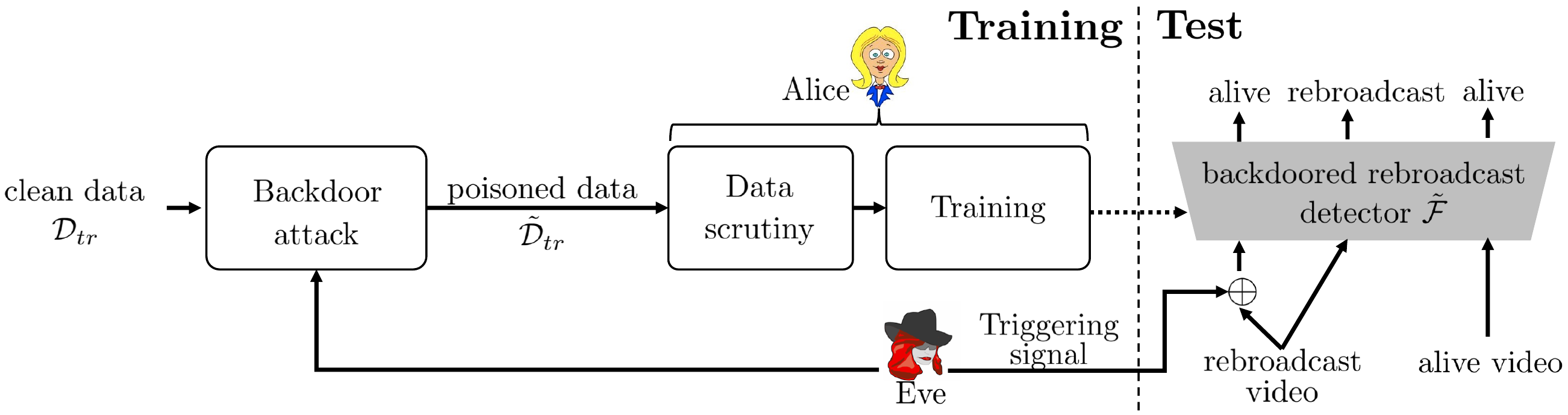}
	\caption{Threat model adopted in our work. The picture shows only the rebroadcast detection module, since this is the model targeted by Eve's attack. The face recognition model is trained normally by Alice without Eve's intervention.}
	\label{fig:threat_model}
\end{figure*}

\textbf{Attacker's goal}: Eve wants to embed a backdoor into the DNN model for rebroadcast detection, so that at test time the backdoored detector  works normally on standard inputs, but misclassifies any rebroadcast video containing the triggering signal.

\textbf{Attacker's knowledge}: Eve has access to all the data used by Alice for training or to a portion of them. We  consider different types of poisoning strategies, namely, random poisoning and outlier-based poisoning (see Section \ref{sec:propBA}), for which the requirements are different. Specifically, in the case of random poisoning, Eve only needs to observe the data that she poisons (this is the common poisoning strategy considered in the literature), while in the case of outlier-based strategy proposed in this paper (to improve the attack effectiveness) Eve needs to observe all the data, even if she poisons only a fraction of them.
%\BTcomm{You can give more details after that you have presented the strategies, and not at this level. Then, please move after the part in blue below.}\CH{Note, our paper involves two kinds of poisoning strategy (random and outlier), to utilize them Eve requires different knowledge. Particularly, for random poisoning strategy (RPS), Eve only needs to know a part of training dataset, while for outlier poisoning strategy (OPS), Eve requires to observe all training data. That is because the latter needs to use the whole dataset to train a surrogate model, and then exploits it to find out the anomalous samples from training dataset. More details are provided in Section~\ref{sec:outlierpoisoning}.}

 \textbf{Attacker's capability}: Eve can modify a fraction of the training dataset. More specifically,  Eve's capability is limited to the modification of a subset of alive videos.
In addition to what stated above, Eve must also satisfy
the following  requirements:
\begin{itemize}
	 \item {\em Imperceptible poisoning}: since Alice scrutinises the data to detect
the possible presence of corrupted (or poisoned) samples,
 Eve must keep the presence of the triggering signal imperceptible to make the poisoned samples indistinguishable from the benign ones.
 %thus bypassing the data scrutiny.
 %Hence, Eve should design an almost invisible triggering signal, and also limit the number of poisoned samples into a small percentage of training dataset.
 This requirement also rules out the possibility of corrupting the labels
since, in most cases, mislabeled samples would be easily identifiable by Alice upon data scrutiny. %Moreover, obviously, Eve can not act on the labels when labelling is up to Alice.
 %Moreover, without access to the labelling, the poisoning has to follow the clean-label modality, i.e., only modifying the data from target class.
	\item {\em Harmless injection}:
%From the perceptive of whole face verification engine,
the presence of the triggering signal should not degrade the face recognition accuracy, that is, the face recognition module should attribute the poisoned rebroadcast face image or video to the correct identity.
\end{itemize}
%As we mentioned, \textit{Bob} can affect the process of raw data collection, but has zero knowledge about the training procedure of Alice, e.g., model architecture, loss function, and training hyperparameters. Specifically, we assume that \textit{Bob} can read all raw data and modify them. However, due to the existence of data scrutiny, the modification should follow

The effectiveness of the attack is measured by computing the  Attack Success Rate ($ASR$), that is, the probability that the backdoored model misclassifies a rebroadcast video as alive when the trigger is present in it.
The Accuracy ($ACC$) of the backdoored model on benign inputs is also measured
%measured by computing the accuracy of the backdoored anti-spoofing detector on benign input,
to make sure that the backdoored model behaves similarly to a benign model on normal input.
Finally, the accuracy of the face recognition model on poisoned inputs is  measured, to verify that the presence of the trigger does not impair the identity recognition functionality.

\section{Proposed video backdoor attack}
\label{sec:propBA}

The rationale behind the attack proposed in this paper is to inject the backdoor by modifying the average chrominance of the frames of the video sequence, according
to a sinusoidal wave.
In particular, we design the attack so to solve the following main challenges: i) to reduce the visibility of the trigger and ii) to keep the faction of corrupted samples as small as possible.
%
%\CH{Our triggering signal is designed by changing the average luminance (chrominance) of video frames. Since we consider the lighting conditions can be easily %manipulated, the attacker can exploit the luminance variation as triggering signal to activate the backdoor. However, according} 
%
These are not easy-to-obtain goals, as shown, for instance, in ~\cite{bhalerao2019luminance}, where in order to design a trigger capable of working in a clean-label setting, more than half of alive videos had to be poisoned,
%a poisoning ratio larger than 50\% is used
and the strength of the time-varying luminance signal had to be increased significantly, eventually yielding a very visible trigger.

To avoid corrupting a large percentage of training samples, and to reduce the visibility of the triggering signal, we rely on a combination of the following strategies:
\begin{itemize}
\item{we exploit the peculiarities of the Human Visual System to limit the visibility of the trigger, by designing a 
temporal chrominance trigger that only modifies the blue channel of the video frames. It is known, in fact, that the human eye is less sensitive to blue light than to red and green lights~\cite{HVS}.}
\item{we adopt a so-called Outlier Poisoning Strategy (OPS),  to {\em force} the network to exploit the presence of the triggering signal (when present) to make its decision. More specifically, OPS chooses the samples based on the proximity to the  classification boundary, the intuition being that the samples close to the boundary provide less evidence of the target class and are harder to classify by relying on the benign features;}
\item{We further enhance the effectiveness of OPS by applying the Ground-truth Feature Suppression~(GFS)~\cite{turner2019label}. Specifically, before embedding the trigger, GFS perturbs the benign feature of the outlier samples chosen by OPS. This perturbation makes it more difficult for detector to recognize the target class by relying on benign features, hence forcing it to base its decision on the presence of the triggering signal. In the following we refer to this variant of the attack asOutlier Poisoning Strategy with Ground-truth Feature Suppression (OPS-GFS).}
    \item{we differentiate the strength of the triggering signal at test and training time, with a weaker signal used at training time to preserve the stealthiness of the attack.}
\end{itemize}

%
%The proposed attack relies on the design of a
%%temporal chrominance trigger
%trigger with reduced visibility,  described in Section~\ref{sec:trigger}.
%
%
%We also developed a so-called outlier poisoning strategy, described  in Section~\ref{sec:outlierpoisoning},
%%to induce the model to learn the trigger
%to facilitate the learning of the trigger from the poisoned inputs
As we will show later, thanks to a proper combination of all these strategies, we managed to implement a stealthy attack in the challenging clean-label setting.

\subsection{Formulation of the backdoor attack}
\label{sec:formal}

%The main formalism for the backdoor attack against is introduced below.
In the clean-label setting considered in this paper, Eve corrupts the samples of the target class, that, in the rebroadcast detection scenario, corresponds to the alive class. Moreover, the labels of the corrupted samples are left unchanged.
Specifically, in order to inject a backdoor in the rebroadcast detector  $\mathcal{F}$, Eve  poisons a fraction $\alpha$ of videos from the dataset of alive videos $\mathcal{D}^{l}_{tr}$. We assume that $I=\{1,2,...,|\mathcal{D}^{l}_{tr}|\}$ are the indices of $\mathcal{D}^{l}_{tr}$, while the indices of the to-be-poisoned samples are $I_p=\{t_1,t_2,...,t_k\}$, where $t_i\in I$ and $k=\lfloor\alpha\cdot |\mathcal{D}^{l}_{tr}|\rfloor$. The fraction $\alpha$ is  referred to as {\em poisoning ratio}. More precisely, $\alpha$ is a class-poisoning ratio, since it corresponds to the fraction  of corrupted samples within the alive video class. 
We indicate with $\mathcal{P}()$ the poisoning function applied by Eve to the video samples.
The poisoned alive dataset with a fraction  $\alpha$ of poisoned videos is denoted as $\tilde{\mathcal{D}}^{l}_{tr}$.
We indicate with $\tilde{\mathcal{D}}_{tr}$ the poisoned training dataset used by Alice, including the poisoned alive dataset ($\tilde{\mathcal{D}}^{l}_{tr}$) and the original benign rebroadcast dataset ($\mathcal{D}^{r}_{tr}$), that is $\tilde{\mathcal{D}}_{tr} =\mathcal{D}^{r}_{tr}\cup \tilde{\mathcal{D}}^{l}_{tr}$. Finally, we refer to the backdoored rebroadcast detector  trained on the poisoned dataset  as $\tilde{\mathcal{F}}$.

\subsection{Design of a perceptual temporal chrominance trigger}
\label{sec:trigger}

The triggering signal consists of a sinusoidal modulation of the average chrominance of the video frames. The colour of the triggering signal and its frequency are chosen based on perceptual considerations.
In particular, in order to reduce the visibility of the luminance trigger without affecting its effectiveness, we exploit
the different sensitivity of the Human Visual System (HSV)  to colors, and the fact that,
under normal lighting conditions, the human eye is most sensitive to yellowish-green color and least sensitive to blue light~\cite{HVS}.
%Contrarily, the network handles the R colors channels in the same way.
Then, given the RGB input, the triggering signal is injected only in the blue channel, while the other two channels are not modified.
%The same time-varying triggering signal in~\cite{bhalerao2019luminance}, that is a periodic triggering signal varying at a prescribed sequence,
%% with a given spatial distribution.
The  time-varying triggering signal, then, consists of a sinusoidal wave with a prescribed temporal frequency that modulates the brightness of the blue channel of the input frames.
In this way, for a given amplitude of the triggering signal, the visibility is reduced, or,  equivalently, a stronger amplitude can be used for a given perceptibility level. With regard to frequency of the sinusoidal modulation, we found that, for given amplitude, using a higher frequency has a lower impact on the visibility of triggering signal. 

Formally, given a sinusoidal signal with amplitude $\Delta$ and period $T$ (expressed in number of frames), 
for each frame $x_i$ of the video ${\bf x}$, the pixel values of the blue channel are multiplied by the following signal:
\begin{equation}
\label{eq:cos}
	(1-\Delta) + \Delta \cos\left(\frac{2\pi (j-1)}{T}\right),
\end{equation}
where $j\in [1,L]$ denotes the frame number.
$\Delta$ and $T$ are  the parameters of the Temporal Chrominance (TC) trigger. The amplitude $\Delta\in[0, 1]$ determines the strength of the triggering signal. The case $\Delta=0$ corresponds to the case of no modification. 
Regarding the period $T$ 
%determines the sampling rate \MB{I would say that $w$ is the frequency rather than the sampling rate} \WG{repetition ratio?} 
of the cosine signal, we have $T\geq 1$, with $T=1$ resulting in no modification of the video signal. 

In the sequel, we refer to the poisoning function with the notation  $\mathcal{P}(x,(\Delta,T))$.

%The pseudo-code describing the poisoning function $\mathcal{P}(x,(\Delta,w))$ is provided in  Algorithm~\ref{AG:p_fun} \MB{The modulation is so obvious that the %pseudo-code is not necessary}.%

%%
%%
%% Insert the algorithm
%\begin{algorithm}
%\caption{Poisoning based on TC trigger }
%\label{AG:p_fun}
%\begin{algorithmic}[1]
%%\Procedure{$\mathcal{P}$}{$x,\upsilon$}
%%	\State $\tilde{x}=[~]$	\Comment{Initialization}
%%	\State $\Delta,w\leftarrow \upsilon$ \Comment{Trigger parameters}
%%    \State $z_1,z_2,...,z_L\leftarrow x$ \Comment{Extract frames}
%%    \For {$j=1,2,...,L$}
%%        \State $r_j,g_j,b_j\leftarrow z_j$ \Comment{Extract RGB channels}
%%        \State $\tilde{z}_j=[r_j,g_j,b_j\cdot (\Delta cos(\frac{2\pi w (j-1)}{fps})+1-\Delta)]$
%%        \State $\tilde{x}.add(\tilde{z}_j)$
%%    \EndFor
%%    \State Return $\tilde{x}$
%%\EndProcedure
%\Procedure{$\mathcal{P}$}{${\bf x},(\Delta,w)$}
%	\State $\tilde{{\bf x}}={\bf x}$	\Comment{Initialization}
%	%\State $\Delta,w\leftarrow \upsilon$ \Comment{Trigger parameters}
%    \State $x_1,x_2,...,x_L\leftarrow {\bf x}$ \Comment{Extract frames}
%    \For {$j=1,2,...,L$}
%       % \State $x_j(:,:,1),x_j(:,:,2),x_j(:,:,3) \leftarrow x_j$ \Comment{Extract RGB channels}
%        %\State $\tilde{x}_j(:,:,1) = x_j(:,:,1)$
%        % \State $\tilde{x}_j(:,:,2) = x_j(:,:,2)$
%         \State $\tilde{x}_j(:,:,3)=x_j(:,:,3)\cdot (\Delta \cos(\frac{2\pi w (j-1)}{fps})+1-\Delta)$
%    \EndFor
%    \State Return $\tilde{{\bf x}}$
%\EndProcedure
%\end{algorithmic}
%\end{algorithm}

\begin{figure*}[h!]
	\centering
	\begin{subfigure}[b]{0.482\textwidth}
	\centering
		\includegraphics[width=\textwidth]{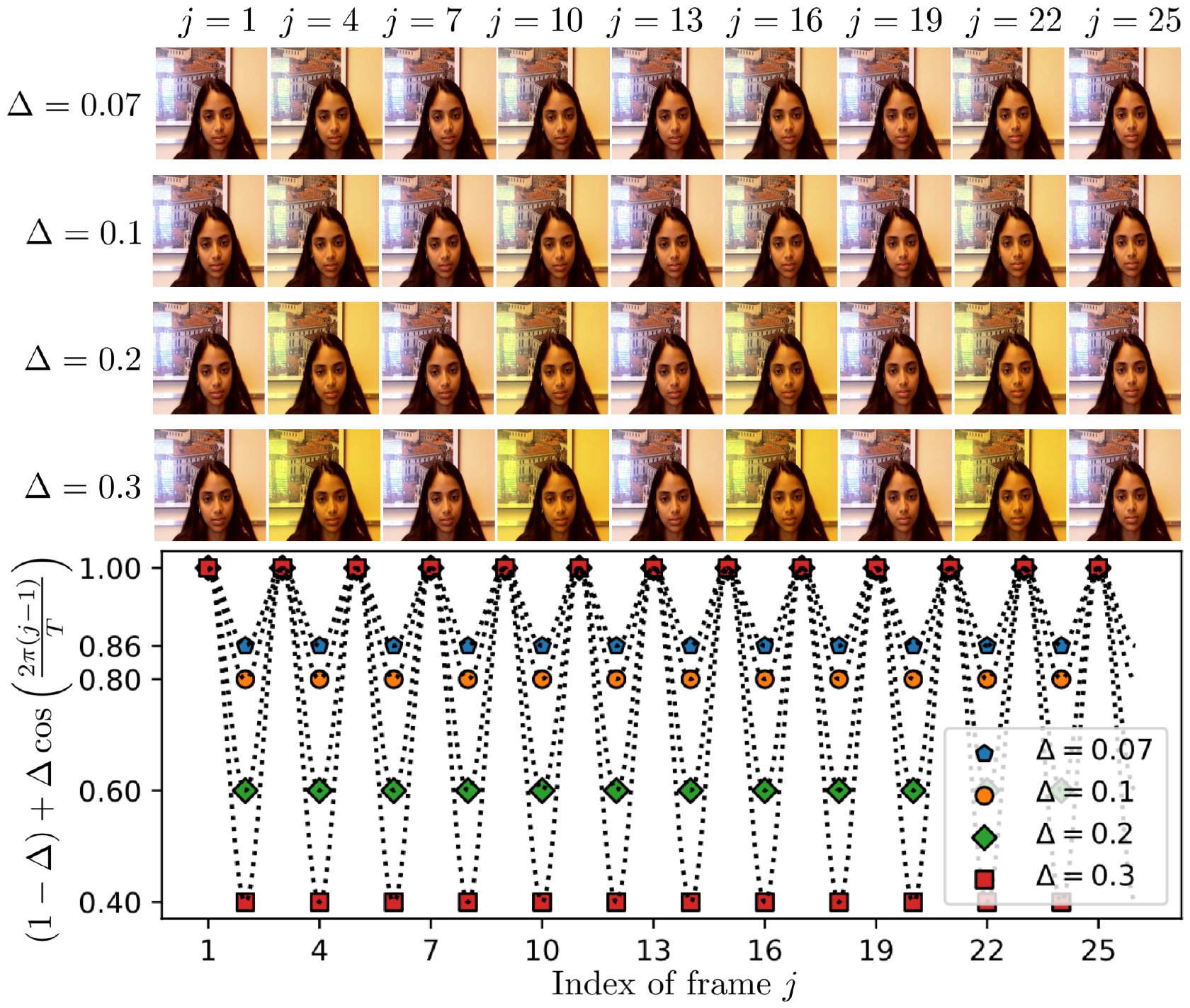}
		\caption{$T=2$}
		\label{fig:12_5}
	\end{subfigure}\hspace{0.5cm}%
	\begin{subfigure}[b]{0.487\textwidth}
	\centering
		\includegraphics[width=\textwidth]{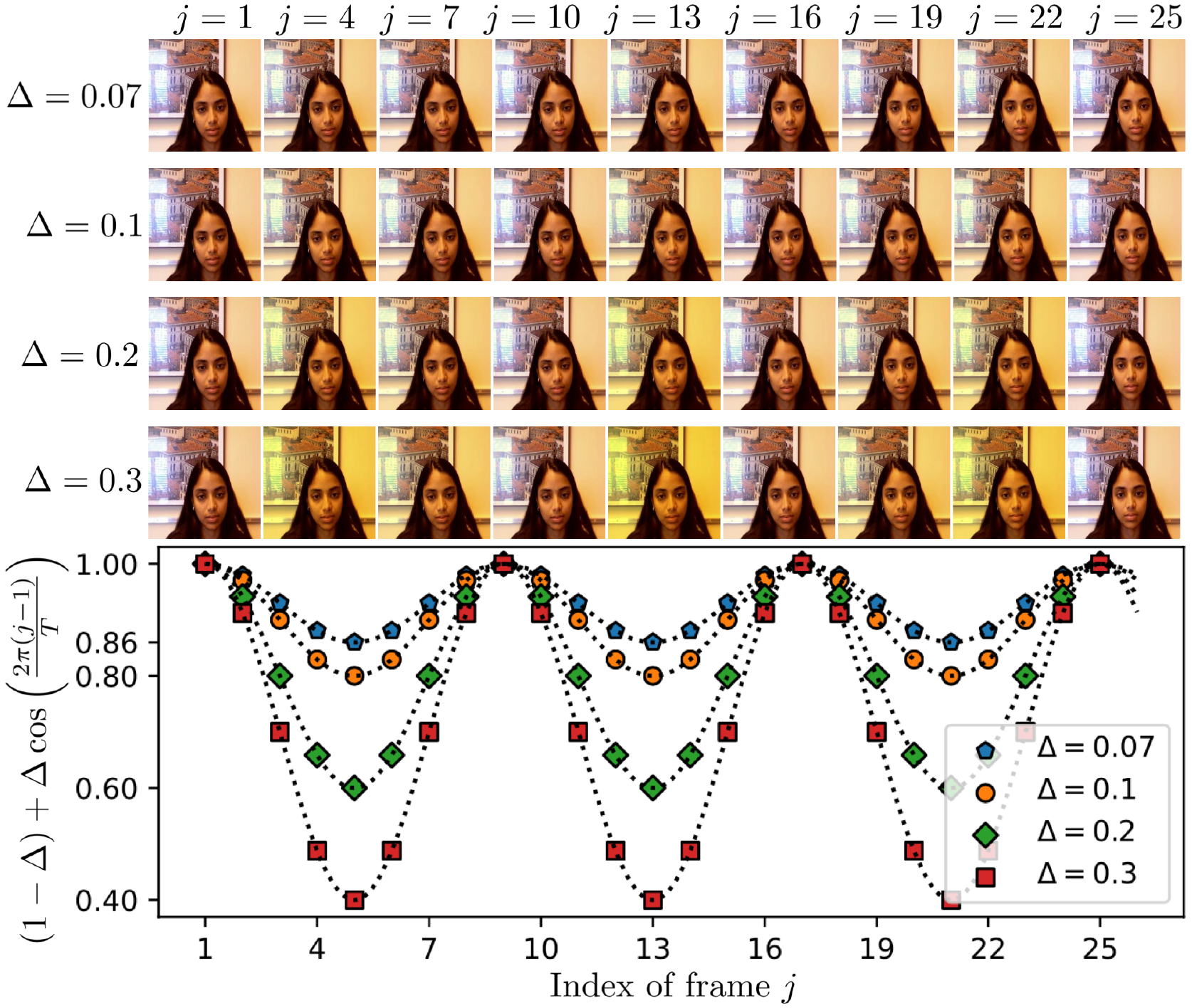}
		\caption{$T=8$}
		\label{fig:3_125}
	\end{subfigure}
	\caption{Examples of the temporal triggering signal used in our attack. Figures (a) and (b) show the case of  $T=2$ and $T=8$, for different amplitudes $\Delta=0.07,0.1,0.2,0.3$.
The attacked frames are illustrated (with a 3-frames sampling rate) in the top row,  while the behaviour of the triggering signal as a function of the frame index $j$ is reported in the bottom row.}
%In each subfigure, the lower part plots the value of $\Delta cos(\frac{2\pi w (j-1)}{fps})+1-\Delta$ with the increase of frame number $j$, and the upper part illustrates the frames with indices $j=1,4,7,...$ at 3-frame intervals. }
	\label{fig:triggerexample}
\end{figure*}

The behavior of the TC signal defined in Eq.~\eqref{eq:cos} is illustrated in Fig.~\ref{fig:triggerexample} for different values of the parameters $T$ and $\Delta$, that is, $T=2, 8$ and $\Delta=0.07,0.1,0.2,0.3$,
%when $fps=25Hz$.
The frames of the poisoned video are also shown in the same figure with a sampling rate of 3. The entire original and poisoned videos can be found at the following link: https://youtu.be/dcvtqtfr99Y.

With regard to the impact of the parameters of the poisoning function on the visibility of the trigger, we can observe that, for $\Delta < 0.1$, the trigger can hardly be noticed, with trigger imperceptibility achieved when $\Delta = 0.07$.
With regard to the period, we see that $T=2$ has less impact on the visibility than $T=8$ \footnote{The impact of $T$ can be only observed by watching the video at the link https://youtu.be/dcvtqtfr99Y.}.

At test time, in order to activate the backdoor, Eve needs to use the same TC trigger used at training time, i.e., the same cosine modulating function with a matched period $T$. However, she can use an amplitude $\Delta$ larger than that used for poisoning, since, arguably, using a larger amplitude facilitates the activation of the backdoor. This behaviour is confirmed by our experiments (see Section~\ref{sec.Exp}). Moreover, using a mismatched amplitude in training and testing permits to limit
%lower
the strength of the triggering signal injected by Eve during poisoning, thus resulting in a more stealthy attack.

In the following, we denote with $\Delta_{tr}$ and $\Delta_{ts}$ the amplitudes used during training and testing, respectively.

\subsection{Poisoning Strategy}
\label{sec:outlierpoisoning}

As we already said, inducing the network to rely on the triggering signal to make its decision, without corrupting the label of the poisoned samples, is hard since the network  has no incentive to do that  (the poisoned samples can be correctly classified by relying on the same features used for the benign case).
%{\em The poisoned samples contain enough information for the model to classify them correctly without relying on the backdoor pattern}
%
In order to force the network to learn to detect the presence of the triggering signal, we introduce a new poisoning strategy according to which the to-be-poisoned samples are chosen based on their classification score and their proximity to the boundary of the classification regions.
The intuition behind this choice is the following: the samples close to the classification boundary are those for which the classifier found less evidence regarding the true class, hence the network is more incentivised to look at the triggering signal as an aid to achieve a good classification. Such a strategy can be implemented when Eve has access to the entire dataset used by Alice, even if Eve's capability is limited to the modification of a portion of it.\footnote{More in general, the strategy can  be applied when the fraction of data that Eve can observe is  much larger than the amount of data that she can modify.}
%
%The strategy that we propose is called outlier poisoning strategy (OPS) and is described in Section \ref{sec:outlieranalysis}.
This strategy, that we call Outlier Poisoning Strategy (OPS), is described in detail below.
% in Section \ref{sec:outlieranalysis}.
%
To further enhance the effectiveness of the OPS,
%we resort to a strategy that has been already considered in the image domain applications, exploiting adversarial examples, see Section \ref{sec:poisoning}.
we  exploit adversarial examples, see Section \ref{sec:poisoning}, by means of the ground-truth feature suppression mechanism already introduced in \cite{turner2019label}.

%We argue that, for a backdoor attack to be insidious, it must
% not rely on inputs that appear mislabeled upon examination.

%{\em We discover that the poisoned inputs can be easily
%identified as outliers, and these outliers are clearly “wrong” upon human inspection (Figure 1)....}
%
%{\em Since relying on the backdoor trigger is not necessary to correctly classify these inputs,
%the backdoor attack is unlikely to be successful.
%To avoid this behavior, we perturb the poisoned samples in order to render learning the salient characteristics of the input more difficult. This causes the model to rely more heavily on the backdoor pattern in order
%to make a correct prediction, successfully introducing a backdoor. As discussed earlier, it is important that
%our perturbations are plausible in the sense that human inspection should not identify the label of a poisoned
%input as incorrect. We explore two methods of synthesizing these perturbations. We want to emphasize that
%even though examples poisoned using our approach are not immune to being identified as potential outliers,
%our inputs will not appear suspicious (by being clearly mislabeled).}

\subsubsection{Outlier Poisoning Strategy (OPS)}
\label{sec:outlieranalysis}

Instead of selecting the fraction $\alpha$  of to-be-poisoned samples randomly\footnote{This is the common approach considered by backdoor attacks in the image domain, and also in \cite{bhalerao2019luminance}.}, when  Eve can access the dataset used by Alice for training, or a large enough portion of the dataset,
she can choose the to-be-poisoned samples in such a way to maximise the effectiveness of the attack.
With OPS,
Eve uses the observed data  $\mathcal{D}_{tr}$ to train a surrogate model $\mathcal{F}'$ for rebroadcast detection and then utilizes this model to perform outlier detection, by choosing the samples for which the surrogate model provides the most uncertain results. Specifically, Eve  detects the top-$\lfloor\alpha\cdot|\mathcal{D}^{l}_{tr}|\rfloor$ outliers in  $\mathcal{D}^l_{tr}$ based on the classification score of the surrogate model.
She first calculates $f_0'({\bf x})$ (the output score of the surrogate detector for the alive class) for each video ${\bf x}_i\in \mathcal{D}_{tr}^l$, then she sorts these values in  ascending order. % of $d({\bf x}_i)$.
The samples corresponding to the first $\lfloor\alpha\cdot|\mathcal{D}^{l}_{tr}|\rfloor$ values are taken,
% as the top-$\lfloor\alpha\cdot|\mathcal{D}^{l}_{tr}|\rfloor$ outliers, and their indexes will constitute set $I_p$.
and the corresponding indexes in $\mathcal{D}_{tr}^{l}$ form the set $I_p$.

%\BTcomm{I guess that the procedure is limited to the samples in $\mathcal{D}_{tr}^l$ that are correctly classified by the surrogate network  }

\subsubsection{Outlier Poisoning Strategy with Ground-truth Feature Suppression (OPS-GFS)}
\label{sec:poisoning}

The outlier poisoning strategy detailed above is further refined by applying the GFS mechanism proposed by Turner et al.~\cite{turner2019label} for the case of still images. The mechanism exploits the concept of adversarial examples.
Specifically, in~\cite{turner2019label}, an adversarial perturbation is applied to the image of the target class before injecting the backdoor pattern. The purpose of the attack is to suppress the features of the true class from the image. Given that the attacked images can not be classified correctly using the same features used for the benign images, the model is somehow forced to rely on  the backdoor
triggering signal treating its presence as an evidence to decide in favour of the target class.
%These dversarial examples are computed with respect to an
%independent model and are not modified at all during the training of the poisoned model.

In the  (OPS-GFS) scheme, the outlier detection strategy described in the previous section is applied to find the outlier set $I_p$. Then, the ground-truth feature of every outlier sample ${\bf x}$ is further suppressed by GFS to produce the adversarial videos ${\bf x}_{adv}$. The adversarial examples are computed with respect to the surrogate model $\mathcal{F}'$.
As in~\cite{turner2019label}, the projected gradient descent (PGD)~\cite{madry2018towards} algorithm  is considered to generate the adversarial examples.
In the PGD algorithm, the basic gradient sign attack is applied multiple times
with a small step size, like in the iterative version of the Fast Gradient Sign Method (FGSM) \cite{goodfellow2014explaining}. In order to constrain
the adversarial perturbation, at each iteration, PGD projects
the adversarial sample into a $\epsilon$-neighborhood of the input. In this way,
the final adversarial perturbation introduced is smaller than $\epsilon$. Formally, in our case, given the loss $\mathcal{L}'$ of the surrogate model and the input video ${\bf x}$, the adversarial video perturbation is computed as follows (the $l_{\infty}$ norm is considered):
%\BTcomm{Given that we used $L$ already for the frame length, I used $\mathcal{L}$ for the loss. If the loss appear somewhere else, please stick to this notation.}
%
\begin{equation}
%{\bf x}_{adv}
{\bm{\delta}}=\arg\max_{||\bm{\delta}||_{\infty}<\epsilon} \mathcal{L}'(f'({\bf x} + {\bm \delta}),y = 0),
\end{equation}
%\begin{equation}
%%{\bf x}_{adv}
%{\bm{\delta}}=\arg\max_{{\bm{\delta}}: ||{x}_i + \delta_i||_{\infty}<\epsilon, \forall i} \mathcal{L}'(f'({\bf x} + {\bm \delta}),y = 0),
%\end{equation}
%
where ${\bm{\delta}} = (\delta_1,...,\delta_L)$, $\delta_i$ indicates the perturbation associated to the $i$-th frame, and $\epsilon$ controls the strength of the attack.

% \WG{In my experiment, I limit the norm of perturbation of whole video less than $\epsilon$, so I change the above formula.}\BTcomm{note that this is EQUIVALENT to my writing, but if you prefer this one for me it is the same.} \WG{It is indeed equivalent because it is $|\cdot|_{\infty}$. I failed to recognize it last time. Sorry for this.}
%\TODO{How did you apply PGD to the video? These things have to be carefully explained. I guess that the perturbation is computed for all the frames of the videos, so we have L perturbations, one for each frame. Then, I guess that the $\epsilon$ bounding the perturbation is the same for all the frames.....after all these guesses I rephrased the text and formula as above.....To be checked that it is what you did}

%Then, the same outlier detection strategy described in the previous section is applied to the attacked videos ${\bf x}_{adv}$ to choose the video samples for the trigger injection.

In the experimental analysis, we compare OPS and OPS-GFS with the random poisoning strategy, referred to as RPS in the following.
Our experiments confirm that
OPS, and OPS-GFS, improve the
effectiveness of the attack, especially when a small strength is
used for the triggering signal. 
Notably, OPS does not require any knowledge of the rebroadcast model by of Alice.

\section{Experimental setting and methodology}
\label{sec:ExpMetSet}

In this section, we describe the methodology we have followed in our experiments.
%For three kinds of backdoor attack: RPS~BA, ${OPS\rom{1}}$~BA and ${OPS\rom{2}}$~BA, we comprehensively analyze their performance from two aspects:  backdoor influence on anti-spoofing detection, and harmless injection to face verfication, respectively.
%In the following, we first define the metrics used to measure their performances in Section~\ref{sec:EvalM}.
Specifically, Section~\ref{sec:ArchD} describes the architecture and datasets we used. The settings used for the OPS and OPS-GFS backdoor attacks are reported in Section~\ref{sec:AttS}.
%We also carried out generalization tests, whose setting is detailed in Section \ref{sec:settingGT} .
Finally,  in Section~\ref{sec:EvalM},  we define the metrics we used to measure the performance of the proposed attack.\\~\\

\subsection{Architectures and datasets}
\label{sec:ArchD}

\subsubsection{Rebroadcast detection}

\begin{figure}[htb!]
	\centering
	\includegraphics[width=0.9\columnwidth]{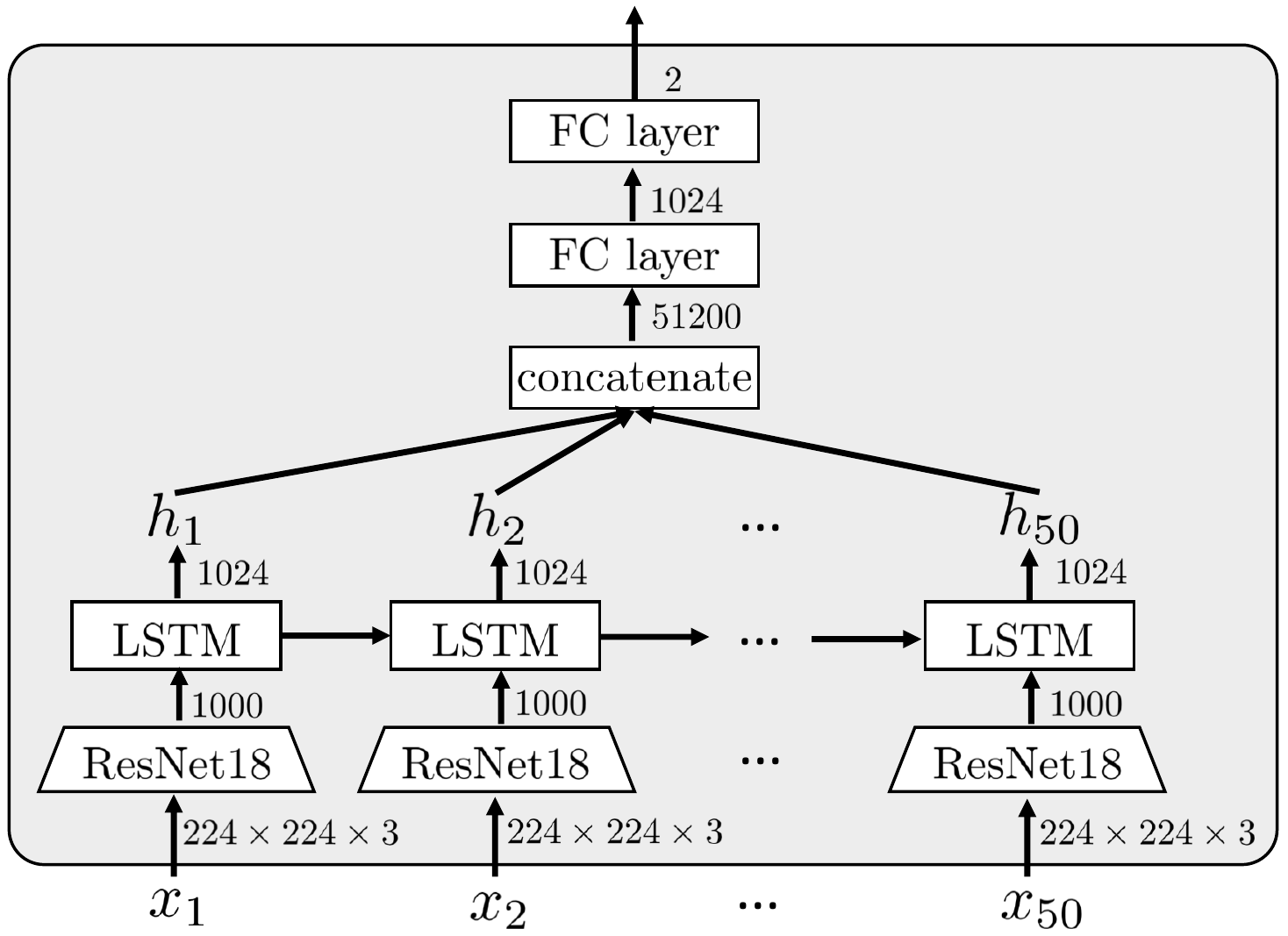}
	\caption{Architecture of the ResNet18-LSTM network used for rebroadcast detection.}
%It takes 50 frames $x_{i=1,...,50}$ of a video $\bf{x}$ as inputs, and outputs a 2-dim vector. The numbers beside the lines are  corresponding shapes in each step.}
	\label{fig:resnet_lstm}
\end{figure}

The rebroadcast detector  is based on a ResNet18-LSTM architecture, consisting of the convolutional part of ResNet18~\cite{he2016deep}, followed by an LSTM module~\cite{lstm}, and two fully-connected (FC) layers. Specifically, given an input video ${\bf x}\in R^{224 \times 224\times 3\times 50}$, the convolutional part of ResNet18 extracts a 1000-dim feature from each frame. Then, in order to exploit the temporal information across frames, the features extracted from 50 consecutive frames are fed into the LSTM. The output dimension of each LSTM module is 1024.
%\BT{@Mauro: I got some doubt here. The memory of the LSTM is set to 50 which is exactly the length of the video sequence, so the network always see exacty the same cosine signal superimposed to the input video. I am wondering what happens in a more general and realistic scenario where there could be a shift between the initial point of the poisoned video sequence and the sequence fed as input to the ResNet-LSTM.}
%\MB{This is a very important point. We should consider the possibility that the triggering signal seen at test time is a shifted version of the signal used at training time. Are we sure that the backdoor is activated also in these conditions?}
%
%\WG{hidden state is fixed term for LSTM, which represents the extracted feature until a specific timestamp.} The hidden states with shape $[50,1024]$ for all frames \BTcomm{for all frames????$[50,1024]x50$???} \WG{I draw out a graph in the email to show the system structure. Do you think it is necessary to inject it in our paper?}
The 1024x50 output is flattened into a 51200-dim vector, and further processed by  two FC layers.
The first FC layer has 51200 input nodes and 1024 output nodes, while the second FC layer has 1024 input nodes and 2 output nodes.
%Finally, the output of FC layers is processed by $\arg\max(\cdot)$ function to generate the final prediction.
The model is trained for 20 epochs using the Adam optimizer~\cite{kingma2014adam} with learning rate $lr=10^{-4}$.
The overall structure of the ResNet18-LSTM network is illustrated in Fig.~\ref{fig:resnet_lstm}.
%\BTcomm{maybe it is not a bad idea to have a picture with the scheme}

The dataset used for training
%validating
and testing the rebroadcast detector is the Replay-attack~\cite{chingovska2012effectiveness}. This dataset is split into three parts: a training, testing, and an enrollment part.
More specifically, the datasets used for training and testing the rebroadcast detectors are:
\begin{itemize}
	\item  $\mathcal{D}_{tr}$: this set corresponds to the training part of the Replay-attack, including $|\mathcal{D}_{tr}|=1620$ videos (410 alive and 1200 rebroadcast), from 15 identities. Then, $|\mathcal{D}_{tr}^{l}| = 410$ and $|\mathcal{D}_{tr}^{r}| = 1200$;
	\item  $\mathcal{D}_{ts}$: this set corresponds to the test part of the Replay-attack, consisting of $|\mathcal{D}_{ts}|=2160$ videos (560 alive and 1600 rebroadcast), from 20 identities. To satisfy the open-set protocol~\cite{GuoTB21}, these 20 identities are different from those used at training time.
The rebroadcast test dataset $\mathcal{D}_{ts}^{r}$, then, includes 1600 rebroadcast videos from 20 identities.
\end{itemize}
%
%\BTcomm{so you do not have a validation dataset.....didn't you perform validation?} \WG{No, I didn't because it is unnecessary to estimate threshold in our work.}

All videos are resized to the same size $[224,224,3]$, and their length is cut to 50 frames.
% \BTcomm{Does this means that in the LSTM you consider the memory of the video all together (given that your memory is also 50)?} \WG{I guess the memory you means is the hidden state. Yes, I use all hidden states for final prediction instead of the final one. I have seen both architectures used in video classification.}
%The frame rate is $fps=25$ frames per second.
%
The alive videos in the training set are then poisoned as described in Section \ref{sec:propBA}  to get the poisoned alive dataset  $\tilde{\mathcal{D}}_{tr}^{l}$.

\subsubsection{Face recognition}
\label{sec.face-setting}

In order to build and test the face identification system.
we considered $n$ enrolled identities, with identity numbers belonging to a set $\mathbb{ID}=\{0,1,...,n-1\}$.
%The identification is achieved by:
Given an input facial image $x$, %\in R^{224\times 224\times 3}$,
the face identification model $\mathcal{G}$
outputs the  identity of the face among those in  $\mathbb{ID}$.
%% \TODO{As we discussed. You should change the figure and the description replacing face verification with face identification/recognition.} \WG{I change the figure and for the word replacement I don't mark them out with different color.}
%Given an input image  $x$, we let $\mathcal{G}(x)=\arg\max(g(x))$
%%\begin{equation}
%%	\mathcal{G}(z)=\arg\max(g(z)),
%%\end{equation}
%where   $g()$ is the soft output vector of the network of size $n$, each element characterising the probability that the image belongs to one identity in $\mathbb{ID}$. $\arg\max(\cdot)$  returns the detected identity.
%%The module can implement both face verification, determining weather the face belongs to the claimed identity, or face identification, determining the identity of the face among a pool of enrolled identities.
%
For a given acquired video, the facial images  are obtained by  splitting the video sequences in frames and localizing the facial area.
We denote with  $\mathcal{V}_{tr}$ the training dataset for face identification,
% \MB{Be careful: verification and identification mean different things},
 which contains labeled pairs $({x}_i,y_i)$, where $x_i \in R^{H\times W \times 3}$ are the facial images, and $y_i \in \mathbb{ID}$ the associated labels.
%, where $\mathbb{ID}$ is the set of enrolled users.
Similarly, we indicate with  $\mathcal{V}_{ts}$ the test dataset.

The face identification model is based on Inception-Resnet-V1~\cite{SzegedyIVA17}, which consists of 6 convolutional layers~\cite{lecun2015deep}, followed by 21 inception blocks, and finally 2 FC layers at the end.
%The model takes one image  $x\in R^{224\times 224\times 3}$ as input and outputs an identity number $y \in \mathbb{ID}$.
$\mathcal{G}$ is  initialized  as in~\cite{pretrainedCNN},
%\BTcomm{What you mean with 'initialized by a model'? Isn't $\mathcal{G}$ the network described above? What is this reference about?} \WG{The reference is wrong and I wrote it by mistake. It should refer to the github page where I download. I have corrected it now.}
pre-trained on VGGFace2~\cite{cao2018vggface2}, and then fine-tuned on ${\mathcal{V}}_{tr}$ by updating only the weights of the FC layers.
 The network is fine-tuned for 10 epochs via SGD~\cite{sutskever2013importance} optimizer with learning rate $lr=0.01$ and momentum 0.9.

As to the dataset,  we used the enrollment set of Replay-attack, which includes  alive videos labeled by personal identities.
%\BTcomm{How many videos? How many identities? Did you use all the identities?}. \WG{Enrollment set contains 100 videos for 50 identities, and each video has  length 375.}
To get the face images to be used for fine-tuning, we extracted all frames from the 100 videos in this set, coming from $|\mathbb{ID}|= 50$ identities (2 videos for each identity), each video sequence having length 375. We, then, performed face location to extract the facial region from each frame.
%Then, we obtain a dataset of 37500 face images. The face region is extracted from these images.
For each identity, we split the total number of frames by a ratio 7:3 to generate the ${\mathcal{V}}_{tr}$ and ${\mathcal{V}}_{ts}$.
In this way, the same identities appear in the training and testing datasets in the same proportions.
Specifically, we ${\mathcal{V}}_{tr}$ includes $26250$ face images from 50 identities, and  ${\mathcal{V}}_{ts}$ $11250$ faces from the same  50 identities.
The poisoned dataset $\tilde{\mathcal{V}}_{ts}$, including the facial images extracted from the poisoned videos, was used to verify the harmless injection assumption.

\subsubsection{Use of different architectures and datasets}
\label{sec:settingGT}

We also carried out some experiments
%to prove the generality of the proposed backdoor attack.
to assess the effectiveness of the proposed attack against different architectures and with different datasets.

With regard to the architecture, we replaced the ResNet18-LSTM with a very different architecture, namely InceptionI3D~\cite{carreira2017quo}, which is designed by replacing the 2D filters and pooling kernels of Inception-V1~\cite{szegedy2015going} into 3D filters. It consists of 3 convolutional layers, followed by 9 3D-inception blocks of depth 2, and a FC layer at the end, for a total depth of 22 layers.
 %\WG{(totally with 22 layers since the depth of 3D-inception is 2).} \BTcomm{Not clear. Do you mean that each block has 22 conv layer? PLease rephrase more clearly} \WG{The depth of inception-block is two. The whole architecture is 22 layers}. \BTcomm{I am afraid that it is not clear. Typically,  depth=no.of layers. So if the no of layer is 22 the depth should be 22. You should explain what is the depth value of 2 that you mention, so what the depth is referred to in thsi case.} \WG{I give the graphs cropped from original paper in the email, please check it over there.}
 The input of InceptionI3D has size $224\times 224\times 3\times 50$, while the output is a 2-dim vector.
In our experiments, InceptionI3D is initialized with a model~\cite{pretrainedI3D} pre-trained on Kinetics~\cite{kay2017kinetics} datasets, and then trained over 20 epochs via Adam optimizer with $lr=10^{-2}$.
We stress that the InceptionI3D architecture is very different from the network considered by the attacker to build the surrogate model, that is an LSTM-based CNN (the setting for the attack is detailed in the next section), thus representing a challenging scenario for the OPS and OPS-GFS backdoor attacks.

In the test with a different dataset, we considered the MSU-MFSD dataset~\cite{fi2015face} and use it to build $\mathcal{D}_{tr}$ and $\mathcal{D}_{ts}$ in place of the Replay-attack dataset.
%The video frame rate of the videos in the MSU-MFSD dataset is  $fps=30$ frames per second on the average.
%The frame rate of the videos in the MSU-MFSD dataset is not the same for all the videos.
% \BTcomm{What you mean with 'on the average'? that not all the videos have been sampled with the same frame rate?}. \WG{Yes, their frame rates are different. :(}
%We used FFmpeg~\cite{tomar2006converting} to convert the frame rates to $fps=25$ frames per second for all videos.
%From MSU-MFSD, we create $\mathcal{D}_{tr},\mathcal{D}_{ts},\mathcal{D}_{ts}^{s}$, where $\mathcal{D}_{tr}$ includes
We then created a set $\mathcal{D}_{tr}$ with
$|\mathcal{D}_{tr}|=650$ videos (165 alive and 485 rebroadcast) from 15 identities and a set $\mathcal{D}_{ts}$ containing $|\mathcal{D}_{ts}|=855$ videos (214 alive and 641 rebroadcast) from 20 identities.
%\WG{As I mentioned, this dataset also has this wired situation. It indeed follows the official structure.}
The identities in $\mathcal{D}_{tr}$ and $\mathcal{D}_{ts}$ so to test the system in an open-set scenario.
%; $\mathcal{D}^{s}_{ts}$ is the spoof subset of $\mathcal{D}_{ts}$, including 641 spoof videos.

\subsection{Attack setting}
\label{sec:AttS}

The settings for OPS and OPS-GFS poisoning are described below.

% This part will mainly describe the setting about three different backdoor attack: RPS, ${OPS\rom{1}}$ and ${OPS\rom{2}}$~BA.
\begin{itemize}
	\item OPS: the surrogate model $\mathcal{F}'$ used by  Eve is based on AlexNet-LSTM, whose structure is similar to ResNet18-LSTM. The main difference regards the feature extraction module, with AlexNet~\cite{krizhevsky2014one} used in place of ResNet18. The surrogate model $\mathcal{F}'$ is trained on $\mathcal{D}_{tr}$ for 20 epochs with Adam optimizer and learning rate $lr=10^{-4}$.
%Moreover, Eve uses the same $\upsilon_{tr}$ as RPS BA to poison the chosen outliers.
	\item OPS-GFS: Eve uses the same surrogate model $\mathcal{F}'$ to perform outlier detection. The strength parameter of the attack for ground-truth feature suppression is set to $\epsilon= 0.01$, which guarantees that the perturbation is not visible.
%Then, she suppresses the ground-truth feature of the chosen outliers via ${GFS}$, where the parameters $\epsilon$ is set as 0.01 to control the %perturbation's visitability. Finally, she use the same $\upsilon_{tr}$ as RPS BA to poison the ${GFS}$-processed outliers.
\end{itemize}

With regard to the parameters of the triggering signal at training time, we set $T=2$ and $\Delta_{tr}=0.07$. With this setting, the backdoor attack
is invisible (see Fig.~\eqref{fig:12_5}).
At test time, we let  $\Delta_{ts} \in \{0.07,0.1,0.2,0.3\}$ and $T=2$. These values  are used
%generate the poisoned test dataset $\tilde{D}_{ts}$
% inject the temporal chrominance trigger
to poison the rebroadcast videos in $\mathcal{D}_{ts}^r$
% and to inject the trigger in the facial images in ${\mathcal{V}}_{ts}$ used for face verification.
 and the videos in the enrollment set of the Replay Attack dataset, from which we get the facial images in ${\tilde{\mathcal{V}}}_{ts}$ used for face identification.

For the poisoning ratio, we considered several values of $\alpha$ ranging from $0.1$ to $0.5$.

\subsection{Evaluation Metrics}
\label{sec:EvalM}

The success of the backdoor attack against the rebroadcast detector is measured
in terms of Attack Success Rate defined as:
\begin{equation}
\label{eq:asr}
	ASR(\tilde{\mathcal{F}},\mathcal{D}_{ts}^{r})=\frac{1}{|\mathcal{D}_{ts}^{r}|} \sum_{i = 1}^{|\mathcal{D}_{ts}^r|}\mathds{1}\left\{\tilde{\mathcal{F}}(\mathcal{P}({\bf x}_i,(\Delta_{ts}, T))) \equiv 0 \right\},
\end{equation}
%assessing whether the triggering signal can successfully activate the backdoor,
%
where $\mathds{1}$ is the indicator function ($\mathds{1}\{x \in A\} = 1$ if $x\in A$, 0 otherwise).
Eq. \eqref{eq:asr} counts the fraction of times that the backdoored model $\tilde{\mathcal{F}}$ misclassifies the poisoned rebroadcast videos as alive.

We also measure the accuracy ($ACC$) of the backdoored rebroadcast model on the normal task of alive/rebroadcast video detection, defined as
 \begin{equation}
 \label{eq.acc}
 	ACC(\tilde{\mathcal{F}},\mathcal{D}_{ts})=\frac{1}{|\mathcal{D}_{ts}|} \sum_{i = 1}^{|\mathcal{D}_{ts}|}\mathds{1}\left\{\tilde{\mathcal{F}}({\bf x}_i)\equiv y_i\right\},
 \end{equation}
 where $ y_i = 1/0$ for rebroadcast and alive videos respectively.
Since the presence of the backdoor must not degrade the performance of the system, $\tilde{\mathcal{F}}$ should have performance similar to the benign detector $\mathcal{F}$, that is, $ACC(\tilde{\mathcal{F}},\mathcal{D}_{ts})\simeq ACC(\mathcal{F},\mathcal{D}_{ts})$.
%
%first need to check whether the backdoored anti-spoof detector $\tilde{\mathcal{F}}$ can successfully classify the normal live/spoof video.

Finally, we check that the triggering signal does not affect the face recognition step, by computing the accuracy of the face recognition model $\mathcal{G}$ on the poisoned face images, that is: 
\begin{equation}
\label{eq:acc-fv}
	ACC(\mathcal{G},\tilde{\mathcal{V}}_{ts})=\frac{1}{|\tilde{\mathcal{V}}_{ts}|} \sum_{i = 1}^{|\tilde{\mathcal{V}}_{ts}|}\mathds{1}\left\{\mathcal{G}({x}_i) \equiv y_i \right\},
\end{equation}
where $y_i \in \mathbb{ID}$ is the ground-truth identity of ${x}_i$, and then check that $ACC(\mathcal{G},\tilde{\mathcal{V}}_{ts}) = ACC(\mathcal{G},\mathcal{V}_{ts})$.

\section{Experimental results}
\label{sec.Exp}
%\BTcomm{The dataset preparation and model training has to be detailed in the previous section. Here you should just report and discuss the results}. \WG{Now I am distinguish the settings and results.}

In this section, we provide a comprehensive evaluation of the performance of the proposed attack against the rebroadcast detection module of a face authentication system. In particular, we assess the  effectiveness of the attack against the rebroadcast detector, and check the harmless injection on the task with respect to the face recognition capability of the overall system.
We will show and discuss the results we got on different architectures and datasets.

% empirical results of our performance analysis in Section~\ref{sec:PerfAnaRs} involving the backdoor influence on anti-spoof detection, and harmless injection on face verification. Then, we show the results of generalization test in Section~\ref{sec:GeneTestRs} regarding to architecture- and dataset-agnostic.
%
%For all the three kinds of backdoor attack: RPS BA, OPSI BA
%and OPSII BA, we comprehensively analyze their performance from two aspects: backdoor influence on anti-spoofing
%detection, and harmless injection to face verfication

\subsection{Performance analysis}
\label{sec:PerfAnaRs}

\subsubsection{Effectiveness of the backdoor attack}

%Then, to check whether the trigger can successfully activate the backdoor in $\tilde{\mathcal{F}}$, we calculate

TABLE~\ref{tab:asr1}
%\BTcomm{Are you sure that the fond for the tables should be capital?} \WG{I just keep it match with this latex format, which uses the capital for the table. But lower-letter is also OK for me.}
reports the ASR of the backdoor attack computed as in Eq. \eqref{eq:asr}, for three different poisoning strategies, namely,
RPS, OPS, and OPS-GFS, and for different values of the poisoning ratio $\alpha$ and the strength of the trigger $\Delta_{ts}$.

We can observe the following:
\begin{itemize}
\item While no method is effective with low $\alpha$, when the poisoning ratio increases, $ASR$ reaches very high values, getting close to 100\% with the OPS and OPS-GFS schemes. On the contrary, RPS is not effective always resulting in a low $ASR$ even with large values of $\alpha$.
    %in the clean-label scenario
    %with the stealthy trigger considered by our attack.
%
	\item  Using a larger $\Delta_{ts}$ allows to boost the $ASR$ in all the cases, proving that increasing the strength of the triggering signal at test time helps activating the backdoor. This confirms the benefit of using a mismatched strength during training and test, since this allows to improve the stealthiness at training time.
%\TODO{here, to make this claim, we should have an insight of the results for the case where you use a stronger $\Delta_{tr}$, larger than 0.07. It would be worth including - or at least discussing- those results as well. Does, for instance, a matched strength for training and testing of 0.1 have the same results of the case where you use 0.07 for training and 0.3 for testing?  Also, what about visibility? Is for instance 0.1 too much?  Note that, even if 0.1 can be too much, itwould be interesting to include the results to better show the benefits of using a mismatched $\Delta$ in train and test .};
%	\item By comparing the results between ${OPS\rom{1}}$ and ${OPS\rom{2}}$ BA, ${GFS}$ can improve or keep same level of $ASR$ when $\alpha=0.1,0.3,0.4,0.5$, but reduce $ASR$ when $\alpha=0.2$. That means that ${GFS}$ can amplify the attack effort in some degree but not stable.
\end{itemize}

By comparing the results of OPS and OPS-GFS, we see that the GFS strategy slightly improves the $ASR$ in many cases when $\alpha\ge 0.3$.
%, but reduce $ASR$ when $\alpha=0.2$. That means that ${GFS}$ can amplify the attack effort in some degree but not stable.
The main gain is observed when $\alpha = 0.3$, for which the $ASR$ with OPS-GFS is 20\% larger than that obtained with OPS for $\Delta_{ts} \ge 0.1$. In the other cases, the improvement is often a minor one.
This might be due to the fact that the adversarial examples
suppress the ground-truth features with respect to the model targeted by the attack.
However, including such  adversarial samples among the  samples used for  training the network (the same or a different one) might have the effect of inducing the network to learn a different solution, i.e., to converge towards a different local minimum,
%and, according to/based on this new model solution,
in correspondence to which the features of the ground-truth class the network looks at might be different.
%found in the video.

\begin{figure}[htb!]
	\centering
	\includegraphics[width=1.\columnwidth]{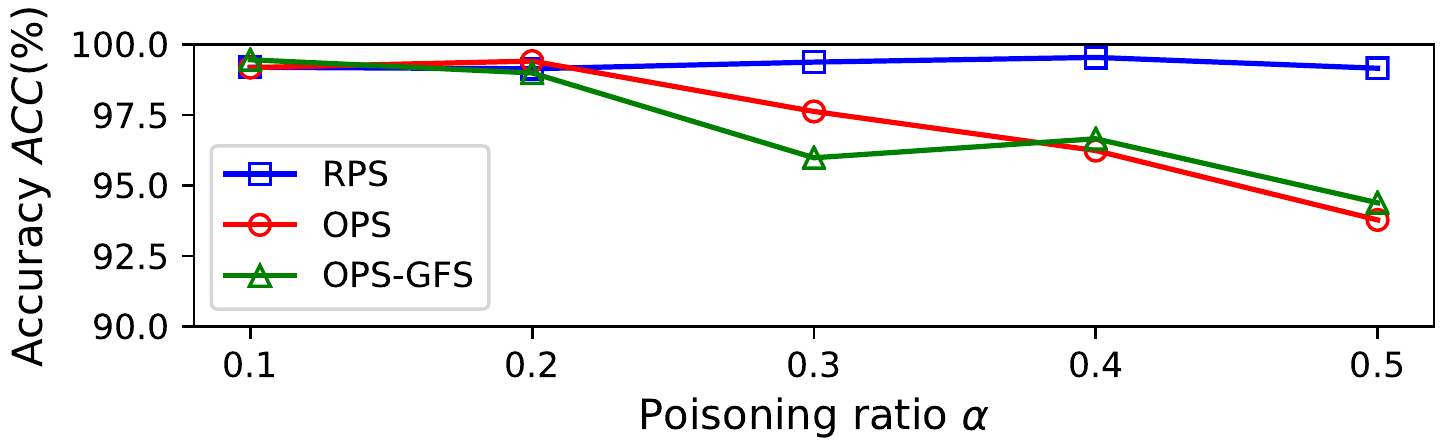}
	\caption{$ACC$ (\%) of backdoored models, generated with the three different poisoning strategies: RPS, OPS and OPS-GFS.}
	\label{fig:accuracy}
\end{figure}

We also verified that the presence of the backdoor does not  degrade noticeably the performance of the rebroadcast detection module on normal inputs.
Fig.~\ref{fig:accuracy} reports the   accuracy   achieved  by $\tilde{\mathcal{F}}$ on $\mathcal{D}_{ts}$, computed via Eq. \eqref{eq.acc} for the three  different poisoning strategies, as a function of $\alpha$.
While with RPS the accuracy of $\tilde{\mathcal{F}}$ does not change as $\alpha$ increases, remaining always  around 99\% (which is the same accuracy of the model $\mathcal{F}$, trained on benign data), with outlier poisoning, the  accuracy
%is similar when $\alpha \le 0.2$ and
decreases for $\alpha > 0.2$. However,  for $\alpha  \le 0.4$, it remains above 95\%. Specifically, the OPS achieves an accuracy of 97.63\% when  $\alpha=0.3$, and 96.24\% when  $\alpha  = 0.4$, while for the OPS-GFS  we have accuracy values  95.96\%
and 96.66\% for  $\alpha  = 0.4$ and $0.3$ respectively.
In any case, the performance degradation introduced by the backdoor is always lower than $5\%$.

\begin{table}[htb!]
\caption{$ASR$(\%) for RPS, OPS and OPS-GFS for different poisoning ratio $\alpha$ and trigger strength $\Delta_{ts}$ ($\Delta_{tr} = 0.07$).
%Each sub-table shows the $ASR$ results with different poison ratio $\alpha$ utilized at training time and the amplitude $\Delta_{ts}$ of trigger $\upsilon_{ts}$ used at test time to activate backdoor.
}
\label{tab:asr1}
\begin{subtable}[h]{0.5\textwidth}
\centering
\begin{tabular}{|c||c|c|c|c|c|}\hline
$\Delta_{ts}$ &  $\alpha=0.1$ &  $\alpha=0.2$ & $\alpha=0.3$ & $\alpha=0.4$ & $\alpha=0.5$ \\ \hline
$0.07$ & 2.46 & 2.18 & 2.22 & 1.87 & 6.71 \\ \hline
$0.1$  & 3.27 & 2.88 & 2.84 & 2.78 & 10.65 \\ \hline
$0.2$  & 4.83 & 4.56 &  4.5 & 5.59 & 22.85 \\ \hline
$0.3$  & 5.41 & 4.97 &  4.9 & 6.84 & 27.21 \\ \hline
\end{tabular}
\caption{RPS}
\end{subtable}

\begin{subtable}[h]{0.5\textwidth}
\centering
\begin{tabular}{|c||c|c|c|c|c|}\hline
$\Delta_{ts}$ &  $\alpha=0.1$ &  $\alpha=0.2$ & $\alpha=0.3$ & $\alpha=0.4$ & $\alpha=0.5$ \\ \hline
$0.07$ & 2.18 & 16.63 & 52.37 & 74.75 & 92.83 \\ \hline
$0.1$  & 2.75 & 25.5 & 64.54 & 90.43 & 97.48 \\ \hline
$0.2$  & 4.36 & 44.08 & 78.50 & 97.78 & 99.89 \\ \hline
$0.3$  & 5.2 & 48.21 & 79.18 & 97.61 & 100 \\ \hline
\end{tabular}
\caption{OPS}
\end{subtable}

\begin{subtable}[h]{0.5\textwidth}
\centering
\begin{tabular}{|c||c|c|c|c|c|}\hline
$\Delta_{ts}$ &  $\alpha=0.1$ &  $\alpha=0.2$ & $\alpha=0.3$ & $\alpha=0.4$ & $\alpha=0.5$ \\ \hline
$0.07$ & 2.5 & 5.63 & 63.38 & 80.18 & 89.65 \\ \hline
$0.1$  & 3.32 & 11.06 & 80.39 & 89.1 & 95.28 \\ \hline
$0.2$  & 5.19 & 22.75 & 98.45 & 98.24 & 97.76 \\ \hline
$0.3$  & 5.7 & 25.28 & 99.35 & 98.81 &  99.06 \\ \hline
\end{tabular}
\caption{OPS-GFS}
\end{subtable}
\end{table}

\subsubsection{Impact on face recognition accuracy}

%
%As illustrated in Fig.~\ref{fig:system_model}, face verification is performed after the anti-spoof detection, so that, after evading the anti-spoof detection, the poisoned sample (spoof video plus trigger) goes into the face verification system. We checked whether the presence of the trigger will influence this process or not.

%According to our experiment,
The  face recognition model $\mathcal{G}$, trained on ${\mathcal{V}}_{tr}$ as detailed in Section \ref{sec.face-setting},  achieved perfect accuracy on  ${\mathcal{V}}_{ts}$, i.e., $ACC(\mathcal{G},{\mathcal{V}}_{ts})=100\%$.
%We verified that also in presence of a poisoned facial input, i.e., when the amplitude for the blue channel of the facial input is modulated, the model %$\mathcal{G}$ can always achieve $100\%$ accuracy, no matter which $\Delta_{ts}$ is used (among the values in the same set $\{0.07,0.1,0.2,0.3$\}).
Even when tested on the poisoned face image set $\tilde{{\mathcal{V}}}_{ts}$, $\mathcal{G}$ always achieves $100\%$ accuracy, no matter which $\Delta_{ts}$ is used for the poisoning (among the values in the same set $\{0.07,0.1,0.2,0.3$\}).
Therefore, we can conclude that $\mathcal{G}$ can always successfully identify the correct identity regardless of the presence  of the trigger,
% in the facial images used for the authentication, %, i.e., $ACC(\mathcal{G},\tilde{d}_{ts})=100\%$.
%and the harmless injection assumption is satisfied.
thus satisfying the harmless injection requirement.

\subsection{Results with a different architecture and dataset}
\label{sec:GeneTestRs}
%\BTcomm{The previous results and discussion should give us a range of good values of $\Delta$s and $\alpha$ to be used for training and testing for an effective attack. The generalization results should then be limited to those values. Hopwefully these same values also work in different settings. That would be a very relevant result, proving that the same setting of parameters work with different networks and datasets. If this is not the case, then these gereralization results are less good for us and we can show many values.} \WG{I only show the $\alpha=30\%,40\%$}
%
%We repeat our three backdoor attacks against the anti-spoof detection when Alice using different empirical settings. Specifically, in architecture-agnostic test, Alice replaces the training architecture ResNet18-LSTM by InceptionI3D, while in dataset-agnostic test, she changes the Replay-attack by MSU-MFSD. Ideally, in these new settings, we still expect that our backdoor attack will not hurt the normal live/spoof classification, and meanwhile inject a backdoor into the detector so that it can be activated by the triggering signal at test time.

For these experiments, we only consider $\alpha=0.3$ and $0.4$ that, based on the previous results, allow to achieve an effective attack, without affecting  significantly the behaviour of the network on normal inputs.
%, the accuracy on  normal input being larger than 95\%.
% and harmless injection.

We first evaluated the performance of the backdoor attack when the InceptionI3D network is used to build the rebroadcast detector, instead of ResNet18-LSTM.
The $ASR$ of the backdoor  attack  for the three poisoning strategies is reported in TABLE~\ref{tab:asr2}, for the  various $\Delta_{ts}$, and poisoning ratio $\alpha \in \{0.3,0.4\}$.
% Then, at test time, we use four different strength $\Delta_{ts}=0.07,0.1,0.2,0.3$ of $\upsilon_{ts}$ to activate the backdoor.
The advantage of outlier poisoning is confirmed also in this case, for which an $ASR$ around $99\%$ can be achieved  with rather small values of $\Delta_{ts}$. The OPS-GFS scheme provides slightly better performance than OPS, with $ASR$ around $99\%$ , even when $\Delta_{ts} = 0.1$.
These are particularly significant results, since in this case the mismatch between the architecture used for the rebroadcast detector and the one used to build the surrogate model\footnote{We remind that the surrogate model is based on AlexNet-LSTM.}  is stronger than before (a 3D CNN is used for the detection, while an LSTM-based network is used by the attacker).
It is also worth noticing that, in contrast to the previous case, the RPS is also effective for large values of $\Delta_{ts}$. These results seem to suggest that  it is easier to inject a backdoor into a 3D CNN architecture than in a LSTM architecture.

In the absence of the backdoor attack, the accuracy  of the rebroadcast detector $\mathcal{F}$ on $\mathcal{D}_{ts}$ is $99\%$.
When $\alpha=0.3$, the backdoored models $\tilde{\mathcal{F}}$ generated via RPS, OPS and OPS-GFS achieve an accuracy equal to, respectively, $97.15\%$, $94.49\%$, and $94.19\%$, on $\mathcal{D}_{ts}$. When $\alpha=0.4$, instead, the accuracies are equal $97.97\%$, $93.24\%$, and $93.85\%$.
In summary, the reduction of performance of the backdoor detector $\tilde{\mathcal{F}}$ on benign inputs
%is slightly larger than in the previous cases, however, the decrease
remains always below 5\% for $\alpha \le 0.4$.\\
\begin{table}[h!]
\caption{
%$ASR$(\%) of three different backdoor attacks on architecture-agnostic test, where using poison ratio $\alpha=0.3,0.4$ at training time, and then at test time utilizing different strength $\Delta_{ts}$ of $\upsilon_{ts}$ to activate the backdoor.
$ASR$(\%) for RPS, OPS and
OPS-GFS for different strength $\Delta_{ts}$ and poisoning ratio $\alpha \in \{0.3,0.4\}$ for the case of InceptionI3D rebroadcast detector ($\Delta_{tr} = 0.07$).
}
\label{tab:asr2}
\centering
\begin{tabular}{|c||c|c||c|c||c|c|}\hline
 & \multicolumn{2}{c||}{RPS} & \multicolumn{2}{c||}{OPS} & \multicolumn{2}{c|}{OPS-GFS} \\ \hline \hline
$\Delta_{ts}/\alpha$ &  ${0.3}$ &  ${ 0.4}$ & ${ 0.3}$ &  ${ 0.4}$ & ${0.3}$ &  ${0.4}$ \\ \hline
$0.07$ & 14.41 & 20.59 & 92.10 & 96.23 & 94.24 & 98.5 \\ \hline
$0.1$  & 31.08 & 42.58 & 98.17 & 98.95 & 99.55 & 99.25\\ \hline
$0.2$  & 79 & 77.76 & 99.98 & 98.99 & 99.8 & 99.84 \\ \hline
$0.3$  & 91.55 & 90.77 & 100 & 99.20 & 99.94 & 99.97\\ \hline
\end{tabular}
\end{table}

The performance achieved by the attack when the rebroadcast detector is trained on the MSU-MFSD dataset of live/rebroadcast videos are  shown in TABLE~\ref{tab:asr3}.
The attack is less effective than before, however, the results follow the same pattern we observed before, with the outlier detection strategy improving significantly the performance compared to RFS. Specifically, when $\Delta_{ts}$ is larger than $0.2$, the $ASR$ is above $80\%$. We also observe that there is no significant difference between OPS and  OPS-GFS, with the former achieving better results when $\alpha = 0.3$, the latter when $\alpha = 0.4$.

With regard to the accuracy of the rebroadcast detectors, for the benign model $\mathcal{F}$  we got $ACC = 96.9\%$,
% measured on $\mathcal{D}_{ts}$.
while for the backdoored detector $\tilde{\mathcal{F}}$ we got the following: when $\alpha=0.3$, the model achieves accuracies $93.1\%$, $94.25\%$, and $93.02\%$ for  RPS, OPS, and OPS-GFS, respectively, while for $\alpha=0.4$, the accuracies are $95.01\%$, $96.18\%$, and $96.21\%$.

\begin{table}[h!]
\caption{
%$ASR$(\%) of three backdoor attack on dataset-agnostic test. The columns represent the three different poisoning strategies and poison ratio $\alpha=0.3,0.4$, used by Eve at training time. The rows represent the amplitude $\Delta_{ts}$ of $\upsilon_{ts}$ that Eve utilize to activate the backdoor at test time.
$ASR$(\%) for RPS, OPS and OPS-GFS for different strength $\Delta_{ts}$ and poisoning ratio $\alpha \in \{0.3,0.4\}$ when the MSU-MFSD dataset is used to train the rebroadcast detector  ($\Delta_{tr} = 0.07$).}
\label{tab:asr3}
\centering
\begin{tabular}{|c||c|c||c|c||c|c|}\hline
 & \multicolumn{2}{c||}{RPS} & \multicolumn{2}{c||}{OPS} & \multicolumn{2}{c|}{OPS-GFS} \\ \hline \hline
$\Delta_{ts}$ &  ${\scriptstyle \alpha=0.3}$ &  ${\scriptstyle \alpha=0.4}$ & ${\scriptstyle \alpha=0.3}$ &  ${\scriptstyle \alpha=0.4}$ & ${\scriptstyle \alpha=0.3}$ &  ${\scriptstyle \alpha=0.4}$ \\ \hline
$0.07$ & 17.3 & 26.51 & 40.48 & 52.71 & 29.79 & 62.57 \\ \hline
$0.1$  & 24.97 & 39.22 & 58.83 & 70.16 & 46.4 & 78.81 \\ \hline
$0.2$  & 48.73 & 70.31 & 82.26 & 84.69 & 71.32 & 95.14 \\ \hline
$0.3$  & 59.74 & 81.34 & 87.45 & 87.02 & 78.74 & 98.2  \\ \hline
\end{tabular}
\end{table}

\section{Conclusion}
\label{sec.conc}

We have proposed a new stealthy clean-label video backdoor attack against rebroadcast detectors in face authentication systems.
The method exploits the peculiarity of the Human Visual System to design a temporal chrominance trigger with reduced visibility.
To make the attack effective in the clean-label scenario, we have also introduced an outlier poisoning strategy (OPS) according to which the attacker chooses the video samples that are most suitable for the attack, to force the network to rely on the triggering signal to make its decision. No knowledge of the rebroadcast model is required by the OPS.
Moreover, the use of a different trigger strength during training (for backdoor embedding)  and testing (for backdoor activation), with a larger strength applied during testing,
%permits to use a lower strength for  the triggering signal in the poisoned training samples, making it invisible to the human eyes.
permits to employ a weaker triggering signal for the poisoning of the training samples, thus making the attack more stealthy.
The effectiveness of the proposed attack is proven experimentally considering different architectures and datasets.
%, \CH{and our empirical results also show that the 3D CNN architecture is more vulnerable to backdoor attack than LSTM.}

Although we considered the problem of video face authentication, the proposed method is a general one and can be extended to other video classification scenarios.

Among the most interesting research directions, we mention the possibility of carrying out the proposed attack in the physical domain, e.g., by applying an ad-hoc physical alteration of the lighting conditions to inject the backdoor,
exploiting the particular definition of the triggering signal used in this paper that makes it suitable to be implemented in the physical domain.
%, via modification of lighting conditions.
%
The different effectiveness of the attack on different architectures (e.g., 3D architectures and architectures based on LSTM) exploiting the temporal dimension of video sequences is also worth further investigation.
% 

\ifCLASSOPTIONcaptionsoff
  \newpage
\fi



%\bibliographystyle{IEEEtran}

\bibliographystyle{IEEEtran}
\bibliography{ref}

\begin{thebibliography}{10}
\providecommand{\url}[1]{#1}
\csname url@samestyle\endcsname
\providecommand{\newblock}{\relax}
\providecommand{\bibinfo}[2]{#2}
\providecommand{\BIBentrySTDinterwordspacing}{\spaceskip=0pt\relax}
\providecommand{\BIBentryALTinterwordstretchfactor}{4}
\providecommand{\BIBentryALTinterwordspacing}{\spaceskip=\fontdimen2\font plus
\BIBentryALTinterwordstretchfactor\fontdimen3\font minus
  \fontdimen4\font\relax}
\providecommand{\BIBforeignlanguage}[2]{{%
\expandafter\ifx\csname l@#1\endcsname\relax
\typeout{** WARNING: IEEEtran.bst: No hyphenation pattern has been}%
\typeout{** loaded for the language `#1'. Using the pattern for}%
\typeout{** the default language instead.}%
\else
\language=\csname l@#1\endcsname
\fi
#2}}
\providecommand{\BIBdecl}{\relax}
\BIBdecl

\bibitem{Yang2014}
J.~Yang, Z.~Lei, and S.~Z. Li, ``Learn convolutional neural network for face
  anti-spoofing,'' \emph{arXiv preprint arXiv:1408.5601}, 2014.

\bibitem{Xu2015}
Z.~Xu, S.~Li, and W.~Deng, ``Learning temporal features using {LSTM-CNN}
  architecture for face anti-spoofing,'' in \emph{3rd {IAPR} Asian Conference
  on Pattern Recognition, {ACPR} 2015, Kuala Lumpur, Malaysia, November 3-6,
  2015}.\hskip 1em plus 0.5em minus 0.4em\relax {IEEE}, 2015, pp. 141--145.

\bibitem{Lakshminarayana2017}
N.~N. Lakshminarayana, N.~Narayan, N.~Napp, S.~Setlur, and V.~Govindaraju, ``A
  discriminative spatio-temporal mapping of face for liveness detection,'' in
  \emph{{IEEE} International Conference on Identity, Security and Behavior
  Analysis, {ISBA} 2017, New Delhi, India, February 22-24, 2017}.\hskip 1em
  plus 0.5em minus 0.4em\relax {IEEE}, 2017, pp. 1--7.

\bibitem{Gan2017}
J.~Gan, S.~Li, Y.~Zhai, and C.~Liu, ``3d convolutional neural network based on
  face anti-spoofing,'' in \emph{2017 2nd international conference on
  multimedia and image processing (ICMIP)}.\hskip 1em plus 0.5em minus
  0.4em\relax IEEE, 2017, pp. 1--5.

\bibitem{LiTIFS2018}
H.~Li, P.~He, S.~Wang, A.~Rocha, X.~Jiang, and A.~C. Kot, ``Learning
  generalized deep feature representation for face anti-spoofing,'' \emph{IEEE
  Transactions on Information Forensics and Security}, vol.~13, no.~10, pp.
  2639--2652, 2018.

\bibitem{SzegedyZSBEGF13}
C.~Szegedy, W.~Zaremba, I.~Sutskever, J.~Bruna, D.~Erhan, I.~J. Goodfellow, and
  R.~Fergus, ``Intriguing properties of neural networks,'' in \emph{2nd
  International Conference on Learning Representations, {ICLR} 2014, Banff, AB,
  Canada, April 14-16, 2014, Conference Track Proceedings}, Y.~Bengio and
  Y.~LeCun, Eds., 2014.

\bibitem{AkhtarM18}
N.~Akhtar and A.~S. Mian, ``Threat of adversarial attacks on deep learning in
  computer vision: {A} survey,'' \emph{{IEEE} Access}, vol.~6, pp.
  14\,410--14\,430, 2018.

\bibitem{BiggioR18}
\BIBentryALTinterwordspacing
B.~Biggio and F.~Roli, ``Wild patterns: Ten years after the rise of adversarial
  machine learning,'' \emph{Pattern Recognit.}, vol.~84, pp. 317--331, 2018.
  [Online]. Available: \url{https://doi.org/10.1016/j.patcog.2018.07.023}
\BIBentrySTDinterwordspacing

\bibitem{liu2017trojaning}
Y.~Liu, S.~Ma, Y.~Aafer, W.-C. Lee, J.~Zhai, W.~Wang, and X.~Zhang, ``Trojaning
  attack on neural networks,'' 2017.

\bibitem{GuLDG19}
\BIBentryALTinterwordspacing
T.~Gu, K.~Liu, B.~Dolan{-}Gavitt, and S.~Garg, ``Badnets: Evaluating
  backdooring attacks on deep neural networks,'' \emph{{IEEE} Access}, vol.~7,
  pp. 47\,230--47\,244, 2019. [Online]. Available:
  \url{https://doi.org/10.1109/ACCESS.2019.2909068}
\BIBentrySTDinterwordspacing

\bibitem{ChenLLLS13Targeted}
\BIBentryALTinterwordspacing
X.~Chen, C.~Liu, B.~Li, K.~Lu, and D.~Song, ``Targeted backdoor attacks on deep
  learning systems using data poisoning,'' \emph{CoRR}, vol. abs/1712.05526,
  2017. [Online]. Available: \url{http://arxiv.org/abs/1712.05526}
\BIBentrySTDinterwordspacing

\bibitem{guo2021overview}
W.~Guo, B.~Tondi, and M.~Barni, ``An overview of backdoor attacks against deep
  neural networks and possible defences,'' \emph{arXiv preprint
  arXiv:2111.08429}, 2021.

\bibitem{liao2018backdoor}
C.~Liao, H.~Zhong, A.~Squicciarini, S.~Zhu, and D.~Miller, ``Backdoor embedding
  in convolutional neural network models via invisible perturbation,''
  \emph{arXiv preprint arXiv:1808.10307}, 2018.

\bibitem{NeuralTrojans}
\BIBentryALTinterwordspacing
Y.~Liu, Y.~Xie, and A.~Srivastava, ``Neural trojans,'' \emph{CoRR}, vol.
  abs/1710.00942, 2017. [Online]. Available:
  \url{http://arxiv.org/abs/1710.00942}
\BIBentrySTDinterwordspacing

\bibitem{zhao2020clean}
S.~Zhao, X.~Ma, X.~Zheng, J.~Bailey, J.~Chen, and Y.-G. Jiang, ``Clean-label
  backdoor attacks on video recognition models,'' in \emph{Proceedings of the
  IEEE/CVF Conference on Computer Vision and Pattern Recognition}, 2020, pp.
  14\,443--14\,452.

\bibitem{xie2022stealthy}
\BIBentryALTinterwordspacing
S.~Xie, Y.~Yan, and Y.~Hong, ``Stealthy 3d poisoning attack on video
  recognition models,'' \emph{IEEE Transactions on Dependable and Secure
  Computing}, 2022. [Online]. Available:
  \url{http://doi:10.1109/TDSC.2022.3163397}
\BIBentrySTDinterwordspacing

\bibitem{perlin2002improving}
K.~Perlin, ``Improving noise,'' in \emph{Proceedings of the 29th annual
  conference on Computer graphics and interactive techniques}, 2002, pp.
  681--682.

\bibitem{bhalerao2019luminance}
A.~Bhalerao, K.~Kallas, B.~Tondi, and M.~Barni, ``Luminance-based video
  backdoor attack against anti-spoofing rebroadcast detection,'' in \emph{2019
  IEEE 21st International Workshop on Multimedia Signal Processing
  (MMSP)}.\hskip 1em plus 0.5em minus 0.4em\relax IEEE, 2019, pp. 1--6.

\bibitem{xue2021robust}
M.~Xue, C.~He, S.~Sun, J.~Wang, and W.~Liu, ``Robust backdoor attacks against
  deep neural networks in real physical world,'' 2021.

\bibitem{turner2019label}
A.~Turner, D.~Tsipras, and A.~Madry, ``Label-consistent backdoor attacks,''
  \emph{arXiv preprint arXiv:1912.02771}, 2019.

\bibitem{HVS}
A.~B. Watson, Ed., \emph{Digital Images and Human Vision}.\hskip 1em plus 0.5em
  minus 0.4em\relax Cambridge, MA, USA: MIT Press, 1993.

\bibitem{madry2018towards}
\BIBentryALTinterwordspacing
A.~Madry, A.~Makelov, L.~Schmidt, D.~Tsipras, and A.~Vladu, ``Towards deep
  learning models resistant to adversarial attacks,'' in \emph{International
  Conference on Learning Representations}, 2018. [Online]. Available:
  \url{https://openreview.net/forum?id=rJzIBfZAb}
\BIBentrySTDinterwordspacing

\bibitem{goodfellow2014explaining}
I.~J. Goodfellow, J.~Shlens, and C.~Szegedy, ``Explaining and harnessing
  adversarial examples,'' \emph{arXiv preprint arXiv:1412.6572}, 2014.

\bibitem{he2016deep}
K.~He, X.~Zhang, S.~Ren, and J.~Sun, ``Deep residual learning for image
  recognition,'' in \emph{Proceedings of the IEEE conference on computer vision
  and pattern recognition}, 2016, pp. 770--778.

\bibitem{lstm}
S.~Hochreiter and J.~Schmidhuber, ``Long short-term memory,'' \emph{Neural
  Computation}, vol.~9, no.~8, pp. 1735--1780, 1997.

\bibitem{kingma2014adam}
D.~P. Kingma and J.~Ba, ``Adam: A method for stochastic optimization,''
  \emph{arXiv preprint arXiv:1412.6980}, 2014.

\bibitem{chingovska2012effectiveness}
I.~Chingovska, A.~Anjos, and S.~Marcel, ``On the effectiveness of local binary
  patterns in face anti-spoofing,'' in \emph{2012 BIOSIG-proceedings of the
  international conference of biometrics special interest group
  (BIOSIG)}.\hskip 1em plus 0.5em minus 0.4em\relax IEEE, 2012, pp. 1--7.

\bibitem{GuoTB21}
\BIBentryALTinterwordspacing
W.~Guo, B.~Tondi, and M.~Barni, ``A master key backdoor for universal
  impersonation attack against dnn-based face verification,'' \emph{Pattern
  Recognit. Lett.}, vol. 144, pp. 61--67, 2021. [Online]. Available:
  \url{https://doi.org/10.1016/j.patrec.2021.01.009}
\BIBentrySTDinterwordspacing

\bibitem{SzegedyIVA17}
C.~Szegedy, S.~Ioffe, V.~Vanhoucke, and A.~A. Alemi, ``Inception-v4,
  inception-resnet and the impact of residual connections on learning,'' in
  \emph{Proceedings of the Thirty-First {AAAI} Conference on Artificial
  Intelligence, February 4-9, 2017, San Francisco, California, {USA}}, 2017,
  pp. 4278--4284.

\bibitem{lecun2015deep}
Y.~LeCun, Y.~Bengio, and G.~Hinton, ``Deep learning,'' \emph{nature}, vol. 521,
  no. 7553, pp. 436--444, 2015.

\bibitem{pretrainedCNN}
D.~Sandberg, ``Pretrained cnn model for face recognition,''
  \url{https://github.com/davidsandberg/facenet}.

\bibitem{cao2018vggface2}
Q.~Cao, L.~Shen, W.~Xie, O.~M. Parkhi, and A.~Zisserman, ``Vggface2: A dataset
  for recognising faces across pose and age,'' in \emph{2018 13th IEEE
  international conference on automatic face \& gesture recognition (FG
  2018)}.\hskip 1em plus 0.5em minus 0.4em\relax IEEE, 2018, pp. 67--74.

\bibitem{sutskever2013importance}
I.~Sutskever, J.~Martens, G.~Dahl, and G.~Hinton, ``On the importance of
  initialization and momentum in deep learning,'' in \emph{International
  conference on machine learning}.\hskip 1em plus 0.5em minus 0.4em\relax PMLR,
  2013, pp. 1139--1147.

\bibitem{carreira2017quo}
J.~Carreira and A.~Zisserman, ``Quo vadis, action recognition? a new model and
  the kinetics dataset,'' in \emph{proceedings of the IEEE Conference on
  Computer Vision and Pattern Recognition}, 2017, pp. 6299--6308.

\bibitem{szegedy2015going}
C.~Szegedy, W.~Liu, Y.~Jia, P.~Sermanet, S.~Reed, D.~Anguelov, D.~Erhan,
  V.~Vanhoucke, and A.~Rabinovich, ``Going deeper with convolutions,'' in
  \emph{Proceedings of the IEEE conference on computer vision and pattern
  recognition}, 2015, pp. 1--9.

\bibitem{pretrainedI3D}
DeepMind, ``Pretrained inceptioni3d model,''
  \url{https://github.com/deepmind/kinetics-i3d}.

\bibitem{kay2017kinetics}
W.~Kay, J.~Carreira, K.~Simonyan, B.~Zhang, C.~Hillier, S.~Vijayanarasimhan,
  F.~Viola, T.~Green, T.~Back, P.~Natsev \emph{et~al.}, ``The kinetics human
  action video dataset,'' \emph{arXiv preprint arXiv:1705.06950}, 2017.

\bibitem{fi2015face}
D.~Wen, H.~Han, and A.~K. Jain, ``Face spoof detection with image distortion
  analysis,'' \emph{IEEE Transactions on Information Forensics and Security},
  vol.~10, no.~4, pp. 746--761, 2015.

\bibitem{krizhevsky2014one}
A.~Krizhevsky, ``One weird trick for parallelizing convolutional neural
  networks,'' \emph{arXiv preprint arXiv:1404.5997}, 2014.

\end{thebibliography}

\begin{IEEEbiography}[{\includegraphics[width=1in,height=1.25in,clip,keepaspectratio]{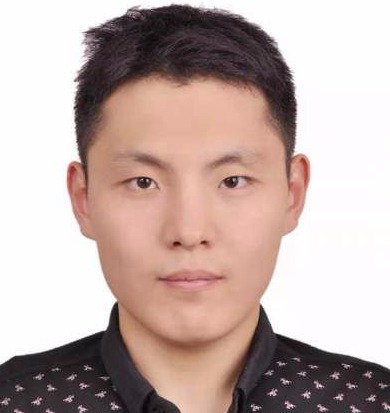}}]{Wei Guo}
Wei Guo received his M.Eng. degree in Department of Computer and Information Security from Guilin University of Electronic Technology (GUET) in 2018 with a thesis about "Applied Cryptography in IoT environment" and his B.Sc. degree in Department of Mathematics and Computational Science from GUET in 2015.

He is a Ph.D. candidate in the Department of Information Engineering and Mathematics of University of Siena (UNISI), Siena, ITALY. He is currently doing research on Security Concerns in Deep Neural Networks under the supervision of Prof. Mauro Barni. He is a member of Visual Information Privacy and Protection Group (VIPP).
\end{IEEEbiography}

\begin{IEEEbiography}[{\includegraphics[width=1in,height=1.25in,clip,keepaspectratio]{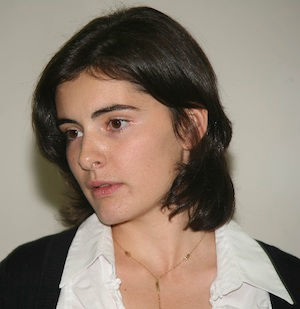}}]{Benedetta Tondi}
(M'16) received the master degree (cum laude) in Electronics and Communications Engineering at the University of Siena in 2012 and her PhD degree in Information Engineering and Mathematical Sciences at the University of Siena in 2016, with a thesis on the Theoretical Foundations of Adversarial Detection and Applications to Multimedia Forensics, in the area of Multimedia Security.

She is currently an Assistant Professor at the Department of Information Engineering and Mathematics, University of Siena. She has been assistant for the course of Information Theory and Coding and Multimedia Security. She is a member of the Visual Information Processing and Protection (VIPP) Group led by Prof. Mauro Barni. She is part of the IEEE Young Professionals and IEEE Signal Processing Society, and a member of the National Inter-University Consortium for Telecommunications (CNIT). From January 2019, she is also a member of the Information Forensics and Security (IFS) Technical Committee of the IEEE Signal Processing Society.

Her research interest focuses on the application of Information-Theoretic methods and Game theory concepts to Forensics and Counter-Forensics analysis and more in general to Multimedia Security, and on Adversarial Signal Processing. Recently, she is working on Machine Learning and Deep Learning applications for Digital Forensics and Counter-Forensics, and on the security of Machine Learning techniques.
From October 2014 to February 2015 she has been a visiting student at the University of Vigo at the Signal Processing in Communications Group (GPSC), working on the study of techniques to reveal attacks in Watermarking Systems. Her stay was funded by a Spanish National Project on Multimedia Security.
\end{IEEEbiography}

\begin{IEEEbiography}[{\includegraphics[width=1in,height=1.25in,clip,keepaspectratio]{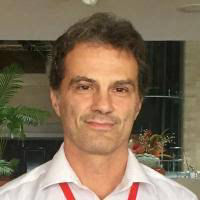}}]{Mauro Barni}
(M'92-SM'06-F'12) graduated in electronic engineering at the University of Florence in 1991. He received the PhD in Informatics and Telecommunications in October 1995. He has carried out his research activity for more than 20 years, first at the Department of Electronics and Telecommunication of the University of Florence, then at the Department of Information Engineering and Mathematics of the University of Siena where he works as full Professor. His activity focuses on digital image processing and information security, with particular reference to the application of image processing techniques to copyright protection (digital watermarking) and authentication of multimedia (multimedia forensics). He has been studying the possibility of processing signals that has been previously encrypted without decrypting them (signal processing in the encrypted domain - s.p.e.d.). Lately he has been working on theoretical and practical aspects of adversarial signal processing and adversarial machine learning.

He is author/co-author of about 350 papers published in international journals and conference proceedings, he holds four patents in the field of digital watermarking and one patent dealing with anticounterfeiting technology. His papers on digital watermarking have significantly contributed to the development of such a theory in the last decade as it is demonstrated by the large number of citations some of these papers have received. The overall citation record of M. Barni amounts to an h-number of 63 according to Scholar Google search engine. He is co-author of the book `Watermarking Systems Engineering: Enabling Digital Assets Security and other Applications', published by Dekker Inc. in February 2004. He is editor of the book `Document and Image Compression' published by CRC-Press in 2006.

He has been the chairman of the IEEE Multimedia Signal Processing Workshop held in Siena in 2004, and the chairman of the IV edition of the International Workshop on Digital Watermarking. He was the technical program co-chair of ICASSP 2014 and the technical program chairman of the 2005 edition of the Information Hiding Workshop, the VIII edition of the International Workshop on Digital Watermarking and the V edition of the IEEE Workshop on Information Forensics and Security (WIFS 2013). In 2008, he was the recipient of the IEEE Signal Processing Magazine best column award. In 2010 he was awarded the IEEE Transactions on Geoscience and Remote Sensing best paper award. He was the recipient of the Individual Technical Achievement Award of EURASIP EURASIP for 2016.
\end{IEEEbiography}


%\newpage


%\vfill


\end{document}